\documentclass[lettersize,journal]{IEEEtran}
\usepackage{amsmath,amsfonts}
\usepackage{algorithmic}
\usepackage{algorithm}
\usepackage{array}
\usepackage[caption=false,font=normalsize,labelfont=sf,textfont=sf]{subfig}
\usepackage{textcomp}
\usepackage{stfloats}
\usepackage{url}
\usepackage{verbatim}
\usepackage{graphicx}
\usepackage{float}
\usepackage{cite}
\usepackage{nicematrix}
\usepackage{colortbl}
\usepackage{threeparttable}
\usepackage[utf8]{inputenc}
\usepackage{newunicodechar}
\newunicodechar{ }{\,}
\begin{document}

\title{\textbf{MSCEKF-MIO: Magnetic-Inertial Odometry Based on Multi-State Constraint Extended Kalman Filter}}

\author{Jiazhu Li, Jian Kuang, Xiaoji Niu ~\IEEEmembership{}
\thanks{}
\thanks{}}

\markboth{}%
{Shell \MakeLowercase{\textit{et al.}}: A Sample Article Using IEEEtran.cls for IEEE Journals}


\maketitle

\begin{abstract}
To overcome the limitations of existing indoor odometry technologies—which often cannot simultaneously meet requirements for accuracy, cost‑effectiveness, and robustness—this paper proposes a novel magnetometer array–aided inertial odometry approach, MSCEKF‑MIO (Multi‑State Constraint Extended Kalman Filter–based Magnetic‑Inertial Odometry). We construct a magnetic field model by fitting measurements from the magnetometer array and then use temporal variations in this model—extracted from continuous observations—to estimate the carrier’s absolute velocity. Furthermore, we implement the MSCEKF framework to fuse observed magnetic field variations with position and attitude estimates from inertial navigation system (INS) integration, thereby enabling autonomous, high‑precision indoor relative positioning. Experimental results demonstrate that the proposed algorithm achieves superior velocity estimation accuracy and horizontal positioning precision relative to state‑of‑the‑art magnetic array–aided INS algorithms (MAINS). On datasets with trajectory lengths of 150–250 m, the proposed method yields an average horizontal position RMSE of approximately 2.5 m. In areas with distinctive magnetic features, the magneto‑inertial odometry achieves a velocity estimation accuracy of 0.07 m/s. Consequently, the proposed method offers a novel positioning solution characterized by low power consumption, cost‑effectiveness, and high reliability in complex indoor environments.
\end{abstract}


\section{Introduction}
\IEEEPARstart{T}{he} indoor positioning market is projected to grow from USD 11.9 billion in 2024 to USD 31.4 billion by 2029, registering a compound annual growth rate (CAGR) of 21.4 $\%$ (MarketsandMarkets, 2024). This growth is mainly driven by the increasing demand for location‑based services (LBS) in smart retail, industrial IoT (IIoT), and emergency response applications. Indoor environments, as quintessential scenarios for LBS, have spurred a wide array of positioning technologies. Infrastructure‑dependent technologies such as pseudolites, Bluetooth, and Wi‑Fi incur high implementation costs and face scalability challenges in large‑scale deployments. Vision‑based localization systems suffer from environmental sensitivity and high power consumption, which fundamentally limits their practical applicability. Magnetic field fingerprinting relies on constructing and maintaining extensive fingerprint databases, resulting in substantial deployment costs. In contrast, inertial navigation system (INS)–based dead reckoning (DR) has emerged as a self‑contained indoor positioning approach, offering autonomous operation and resilience to external signal interference. 
\par
Consumer‑grade devices predominantly employ low‑cost MEMS‑IMUs (micro‑electromechanical systems inertial measurement units), which exhibit limited inherent accuracy and rapid error accumulation when using conventional strapdown navigation algorithms. Therefore, obtaining high‑precision, robustness‑enhanced odometry is essential to constrain error growth and improve the usability and reliability of MEMS‑IMU–based dead reckoning (DR).

\par
Existing odometry approaches include inertial, wheel, visual, and LiDAR-based methods \cite{ref1}. Each modality risks degraded performance under certain operational conditions. Inertial odometry suffers from positioning drift caused by sensor bias accumulation and noise, leading to progressive error growth; wheel odometry is vulnerable to wheel slippage and terrain irregularities, which introduce non‑holonomic constraint violations in motion estimation; visual odometry is highly sensitive to illumination changes and dynamic occlusions, often failing in low‑texture environments; and LiDAR odometry is costly, computationally intensive, and degrades in environments with weak geometric features. No existing odometry method achieves optimal precision, cost‑effectiveness, and robustness simultaneously. Magnetic field odometry offers a lightweight means of measuring velocity. It operates without pre-established magnetic maps, instead leveraging a magnetometer array to provide odometric data—including position and velocity \cite{ref2}.
\par
Due to ferromagnetic interference, indoor magnetic fields exhibit long‑term stability and spatial distinctiveness, making them a reliable positioning source. In 2007, Vissière et al. pioneered the use of indoor magnetic field disturbances to improve IMU‑based position and velocity estimation \cite{ref3}. Vissière et al. derived equations relating magnetic field gradients to user velocity and employed a distributed magnetometer array for velocity estimation. Subsequent studies expanded on this work. In 2011, Dorveaux et al. addressed moving rigid‑body localization by integrating magnetic disturbances with IMU measurements, establishing the Magnetic‑Inertial Navigation (MINAV) framework \cite{ref4}. In 2016, Chesneau et al. integrated inertial sensor and multi‑magnetometer measurements, developing an Extended Kalman Filter (EKF) for data fusion \cite{ref5}. Zmitri and Makia derived higher‑order differential equations of magnetic fields to mitigate noise in gradient measurements, using a distributed magnetometer array to monitor both fields and their spatial derivatives \cite{ref6}. Recent studies have applied a polynomial model to characterize local magnetic fields \cite{ref7}. Further analysis of this model‑based methodology is presented in \cite{ref8}. They augmented the magnetic field model parameters into the INS error‑state vector to facilitate Kalman filter prediction and update processes. Detailed derivations and analyses were subsequently documented in \cite{ref2}, accompanied by real‑world data collection to evaluate algorithm performance. The corresponding experimental datasets and implementation code were also made publicly available.
\par
Recently, a dual‑magnetometer velocity estimation algorithm was implemented for robotic speed determination \cite{ref9}, employing waveform matching between paired vehicle‑mounted magnetometers to compute forward velocity. Experimental results indicate that the approach in \cite{ref9} matches wheel odometry performance in magnetically rich environments. Building on this concept, \cite{ref10} deployed a dual‑magnetometer velocity measurement system on pedestrian helmets and incorporated postural changes into the waveform similarity analysis. However, these methods are restricted to pedestrian forward velocity estimation and exhibit sensitivity to the baseline distance between magnetometers. This paper presents a magnetometer‑array–aided inertial odometry (MIO) framework grounded in the Multi‑State Constraint Extended Kalman Filter (MSCEKF). Compared with the method in \cite{ref2}, the proposed approach achieves superior horizontal velocity and position estimation accuracy and offers improved stability in positioning results.
\par
This work makes the following contributions. First, we derive local magnetic field model equations based on the fundamental assumption that magnetic fields in charge‑free environments exhibit zero divergence and curl. Second, we formulate an error propagation model linking the magnetic field model to relative pose estimation. Third, we incorporate the Multi‑State Constraint Extended Kalman Filter (MSCEKF) in the data fusion phase to enhance robustness. In experiments, we compare the proposed algorithm against MAINS \cite{ref2}, a state‑of‑the‑art open‑source solution. We collected comprehensive field measurements using a custom‑built platform to rigorously evaluate the proposed methodology.
\par
The remainder of this paper is structured as follows. Section II derives the local magnetic field model. Section III details the filtering algorithm for data fusion. Section IV analyzes algorithm performance using both open‑source and proprietary datasets, and examines the impact of varying magnetic gradient intensities. Finally, Section V concludes the paper and proposes directions for future work.

\section{Local Magnetic Field Modeling}
The magnetic field is a three-dimensional vector field whose characteristics are governed by Maxwell's equations. The magnetic field model is denoted by $\boldsymbol{M}(\boldsymbol{r},\boldsymbol{\mu})$, where $\boldsymbol{r} = [r_x, r_y, r_z]^\top$ or $\boldsymbol{r} = [x, y, z]^\top$, represents the position vector from the origin of the local magnetic field model to any arbitrary point in space. The coefficients $\boldsymbol{\mu}$ of the magnetic field polynomial model have a size (number of elements) determined by the polynomial's order. Under the conditions of free space charge and external magnetic interference sources located beyond a 1-meter range, the magnetic field model must satisfy the curl-free and divergence-free assumptions:
\begin{equation} \label{eq1}
    \nabla_{\boldsymbol{r}} \times \boldsymbol{M}(\boldsymbol{r},\boldsymbol{\mu}) = 0
\end{equation}
\begin{equation} \label{eq2}
    \nabla_{\boldsymbol{r}} \cdot \boldsymbol{M}(\boldsymbol{r},\boldsymbol{\mu}) = 0
\end{equation}
Given the curl-free nature of the magnetic field, there necessarily exists a scalar potential function $\boldsymbol{\phi}(\boldsymbol{r},\boldsymbol{\mu})$ satisfying:

\begin{equation} \label{eq3}
    \boldsymbol{M}(\boldsymbol{r},\boldsymbol{\mu}) = \nabla_{\boldsymbol{r}} \boldsymbol{\phi}(\boldsymbol{r},\boldsymbol{\mu})
\end{equation}
In the equation, $\boldsymbol{\phi}(\boldsymbol{r},\boldsymbol{\mu})$ represents an $(l+1)$th order polynomial function of $\boldsymbol{r}$, where $l \in \mathbb{N}$. For the case $l = 1$, the scalar potential function can be expressed as:
\begin{equation} \label{eq4}
    \boldsymbol{\phi}(\boldsymbol{r},\boldsymbol{\mu}) = h(\boldsymbol{r})^\top \boldsymbol{\mu} + c
\end{equation}
\begin{equation} \label{eq5}
    h(\boldsymbol{r})^\top =\left[\begin{array}{lllllllll}
    x & y & z & x y & x z & y z & x^2 & y^2 & z^2
    \end{array}\right]
\end{equation}
\begin{equation} \label{eq6}
    \boldsymbol{\mu} = \left[\begin{array}{lll}
        \mu_1 & \ldots & \mu_9
    \end{array}\right]^\top
\end{equation}
In the equation, $c$ denotes a constant term, while $\boldsymbol{\mu}$ represents the local magnetic field model parameters.
\par
Let $\boldsymbol{\Gamma}(\boldsymbol{r}) = \nabla_{\boldsymbol{r}} h(\boldsymbol{r})^\top$, combining equation \eqref{eq3} and \eqref{eq4} yields:
\begin{equation} \label{eq7}
    \boldsymbol{M}(\boldsymbol{r},\boldsymbol{\mu}) = \nabla_{\boldsymbol{r}} \boldsymbol{\phi}(\boldsymbol{r},\boldsymbol{\mu}) = \boldsymbol{\Gamma}(\boldsymbol{r}) \boldsymbol{\mu}
\end{equation}
where $\boldsymbol{\Gamma}$ is defined as:
\begin{equation} \label{eq8}
    \boldsymbol{\Gamma}(\boldsymbol{r})=\left[\begin{array}{ccccccccc}
    1 & 0 & 0 & y & z & 0 & 2 x & 0 & 0 \\
    0 & 1 & 0 & x & 0 & z & 0 & 2 y & 0 \\
    0 & 0 & 1 & 0 & x & y & 0 & 0 & 2 z
    \end{array}\right]
\end{equation}
Combining with the divergence-free assumption of the local magnetic field in \eqref{eq2}:
\begin{equation} \label{eq9}
    \nabla_{\boldsymbol{r}} \cdot \boldsymbol{\Gamma}(\boldsymbol{r}) \boldsymbol{\mu} = 0
\end{equation}
\begin{equation} \label{eq10}
    \nabla_{\boldsymbol{r}} \cdot \left[\begin{matrix}
        \mu_1 + \mu_4 y + \mu_5 z + 2\mu_7 x \\
        \mu_2 + \mu_4 x + \mu_6 z + 2\mu_8 y \\ 
        \mu_3 + \mu_5 x + \mu_6 y + 2\mu_9 z
    \end{matrix}\right] = 0
\end{equation}
By reorganizing equation \eqref{eq10}:
\begin{equation}  \label{eq11}
    \mu_7 + \mu_8 + \mu_9 = 0
\end{equation}
Let $\mu_9 = -(\mu_7 + \mu_8)$, then the equation can be rewritten as:
\begin{equation} \label{eq12}
    \boldsymbol{M}(\boldsymbol{r}, \boldsymbol{\theta}) = \boldsymbol{\Phi}(\boldsymbol{r})\boldsymbol{\theta}
\end{equation}

Equation \eqref{eq12} represents the derived local magnetic field model. Here, $\boldsymbol{\theta}$ denotes the polynomial coefficients whose dimensionality relates to the model order $l$ through $dim(\boldsymbol{\theta}) = l^2+4l+3$. $\boldsymbol{\Phi}$ is the coefficient matrix. For a 1st-order magnetic field model ($l=1$), $\boldsymbol{\Phi}$ can be expressed as:
\begin{equation} \label{eq13}
    \boldsymbol{\Phi}(\boldsymbol{r}) = \left[\begin{matrix}
    1 & 0 & 0 & y & z & 0 & 2x & 0 \\
    0 & 1 & 0 & x & 0 & z & 0 & 2y \\
    0 & 0 & 1 & 0 & x & y & -2z & -2z
    \end{matrix}\right]
\end{equation}
The local magnetic field model equation shown in Equation \eqref{eq12} is applicable to any coordinate system.
\par
\begin{figure}
    \centering
    \includegraphics[width=0.8\linewidth]{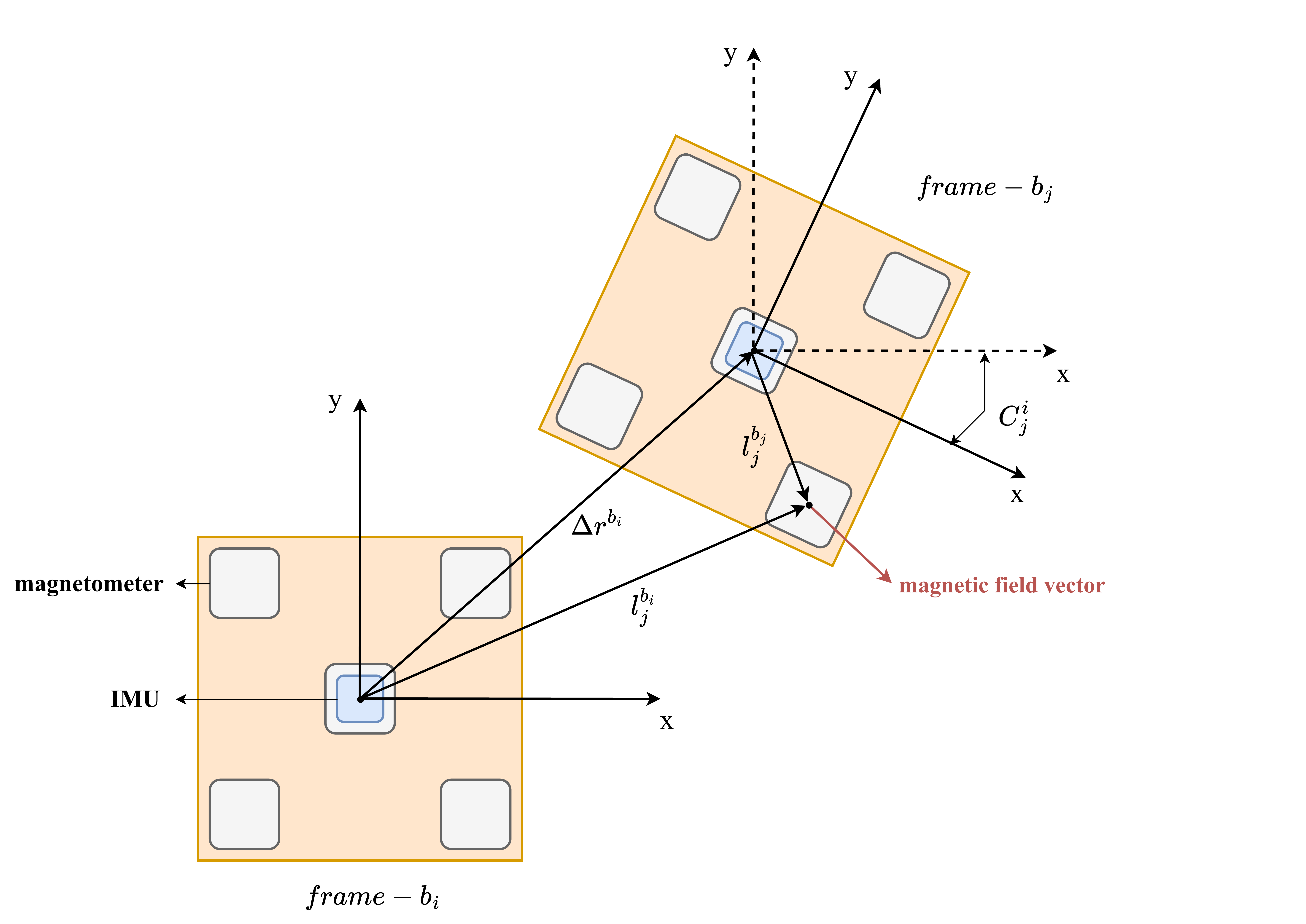}
    \caption{A 2D illustration of the geometric relationship between the body frames at two consecutive times.}
    \label{fig:2d_illus}
\end{figure}

The relationship between magnetic field vectors at the identical spatial location between two consecutive time instants is investigated. The magnetic field vectors in the body frame (b-frame) at two consecutive time instants satisfy the following relationship:
\begin{equation} \label{eq14}
    \boldsymbol{M}^{b_j}(\boldsymbol{r}^{b_j}, \boldsymbol{\theta}_j) = \boldsymbol{C}^{b_j}_{b_i} \boldsymbol{M}^{b_i}(\boldsymbol{r}^{b_i}, \boldsymbol{\theta}_i)
\end{equation}
\begin{equation} \label{eq15}
    \boldsymbol{\Phi}(\boldsymbol{r}^{b_j}) \boldsymbol{\theta}_j = \boldsymbol{C}^{b_j}_{b_i} \boldsymbol{\Phi}(\boldsymbol{r}^{b_i}) \boldsymbol{\theta}_i
\end{equation}
where $b_i$ and $b_j$ denote the body frame at the current and previous time instants, respectively. Fig.\ref{fig:2d_illus} illustrates the geometric relationship between the body frames ($b_i$ and $b_j$) at two consecutive time instants. 

The parameters $\hat{\boldsymbol{\theta}}$ of the magnetic field model can be estimated via least squares using observations from the magnetometer array and Equation \eqref{eq12}.

\begin{equation} \label{eq16}
    \hat{\boldsymbol{\theta}}=(\boldsymbol{\Phi}^\top(\boldsymbol{l}_i^{b_i}) \boldsymbol{\Phi}(\boldsymbol{l}_i^{b_i}))^{-1} \boldsymbol{\Phi}(\boldsymbol{l}_i^{b_i})^\top \boldsymbol{M}^{b_i}
\end{equation}

Here $\boldsymbol{l}_i^{b_i}$ denotes the coordinates of each magnetometer in the $b_i$-frame, $\boldsymbol{M}^{b_i}$ represents the observation vector of each magnetometer, and $\boldsymbol{\Phi}(\boldsymbol{l}_i^{b_i})$ is the coefficient matrix constructed from the coordinate vector $\boldsymbol{l}_i^{b_i}$ following Equation\eqref{eq13}. The estimation covariance matrix for the local magnetic field model parameters $\hat{\boldsymbol{\theta}}$ is:

\begin{equation} \label{eq17}
    \boldsymbol{Cov}_{\hat{\boldsymbol{\theta}}} = \sigma^2(\boldsymbol{\Phi}^\top\boldsymbol{\Phi})^{-1}
\end{equation}
where $\sigma^{2}$ is estimated by calculating the sum of squares of model residuals.

Subsequently, the local magnetic field parameters $\hat{\boldsymbol{\theta}}$ are utilized to predict the magnetic field observation $\hat{\boldsymbol{M}}^{b_i}_{j}$ at the previous time step. We assume that the parameters of the magnetic field model do not change significantly between two adjacent moments: 

\begin{equation} \label{eq18}
    \hat{\boldsymbol{M}}^{b_i}_{j} = \boldsymbol{\Phi}(\boldsymbol{l}_j^{b_i}) \hat{\boldsymbol{\theta}}
\end{equation}
where $\boldsymbol{l}_j^{b_i}$ denotes the coordinates of the magnetometer in the $b_i$-frame at the previous time. Since the magnetometers are rigidly mounted on the platform assembly, the coordinate relationship $\boldsymbol{l}_j^{b_j} = \boldsymbol{l}_i^{b_i}$.

\begin{equation} \label{eq19}
    \boldsymbol{l}_j^{b_i} = \boldsymbol{C}_{j}^{i} \boldsymbol{l}_j^{b_j} + \Delta\boldsymbol{r}^{b_i}
\end{equation}
\begin{equation} \label{eq20}
    \boldsymbol{C}_{j}^{i} = \boldsymbol{C}_{n}^{b_i} \boldsymbol{C}_{b_j}^{n}
\end{equation}
\begin{equation} \label{eq21}
    \Delta\boldsymbol{r}^{b_i} = \boldsymbol{C}_{n}^{b_i} \Delta\boldsymbol{r}^{n}_i = \boldsymbol{C}_{n}^{b_i}(\boldsymbol{r}_{j}^{n} - \boldsymbol{r}_i^{n})
\end{equation}
where $\boldsymbol{C}_{j}^{i}$ represents the relative rotation between the $b_i$-frame and $b_j$-frame at consecutive time steps, $\Delta\boldsymbol{r}^{b_i}$ denotes the projection of the position variation in the $n$-frame onto the $b_i$-frame. Synthesizing Equations \eqref{eq14}, \eqref{eq15}, and \eqref{eq18} yields the predicted magnetic field value $\hat{\boldsymbol{M}}^{b_j}_{j}$ from the previous time step:

\begin{equation} \label{eq22}
    \hat{\boldsymbol{M}}^{b_j}_{j} = \boldsymbol{C}_{j}^{i} \hat{\boldsymbol{M}}^{b_i}_{j} = \boldsymbol{C}_{j}^{i}  \boldsymbol{\Phi}(\boldsymbol{l}_j^{b_i}) \hat{\boldsymbol{\theta}}
\end{equation}

Equation \eqref{eq22} encodes the principle that, within a sufficiently small spatiotemporal neighborhood, the local magnetic field parameters $\boldsymbol{\theta}$ remain constant. Within the model’s effective domain, we attribute observed magnetic fluctuations to changes in the platform’s pose. By collecting magnetic field observations within this domain, we employ perturbation analysis to derive relationships between variations in the magnetic gradient and changes in the platform’s pose. These relationships yield constraint terms derived from observations, which we integrate into the filtering framework to improve state estimation accuracy.

\section{Magnetic-Inertial Odometry with Magnetometer Array Constraint}
\begin{figure*}[ht]
    \centering
    \includegraphics[scale=0.17]{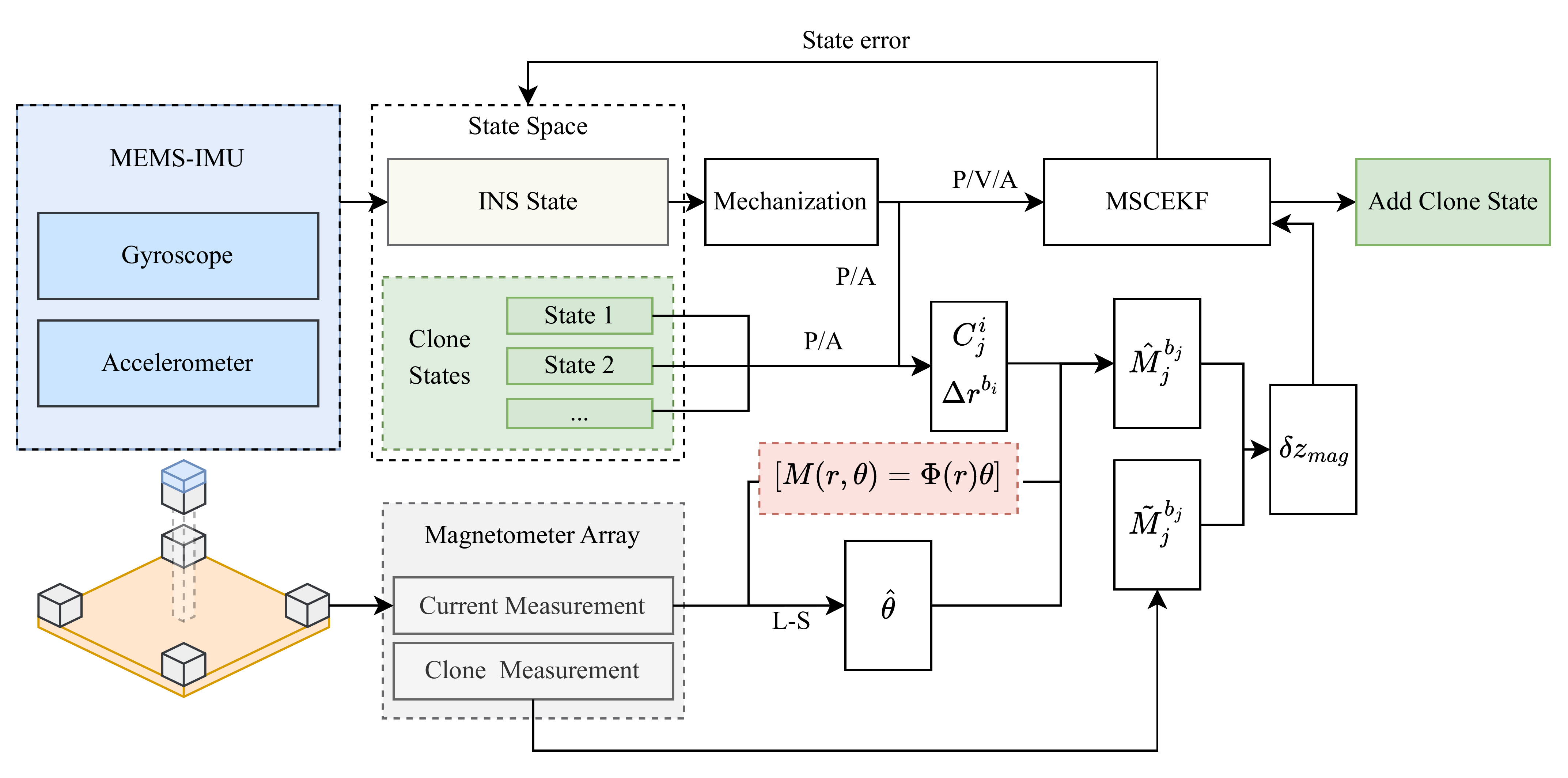}
    \caption{Algorithm flow chart of MSCEKF-MIO}
    \label{fig:mio_flow_chart}
\end{figure*}

This section presents an overview of the proposed data fusion algorithm. Figure \ref{fig:mio_flow_chart} presents the algorithm’s flowchart. First, strap-down inertial navigation mechanization is executed to determine the vehicle’s current position, velocity, and attitude. Next, magnetic array measurements are used to estimate the parameters of the local magnetic field model. Subsequently, historical state information is employed to compute relative pose changes between successive epochs and to predict the corresponding magnetic field models. These predictions are then combined with magnetic array measurements from adjacent epochs to formulate update constraints. Finally, the Multi-State Constraint Extended Kalman Filter (MSCEKF) is applied to fuse magnetic field model variations—observed by the magnetometer array—with relative poses derived from inertial navigation integration, thereby achieving continuous and reliable state estimation. Additionally, the algorithm incorporates magnetic vector constraints in the navigation frame to improve heading angle estimation.

\subsection{Inertial Navigation Algorithm}
The INS Mechanization algorithm is based on the idea that the current position, velocity, and attitude of a moving object can be obtained by integrating acceleration twice and angular rate once, given the initial navigation state. In this paper, a low-cost MEMS-IMU is employed for mechanization, allowing minor corrections (e.g., Earth rotation effects) to be neglected. The simplified mechanization in the navigation frame (n-frame) is expressed as follows:
\begin{equation} \label{eq23}
    \left[\begin{array}{c}
    \boldsymbol{r}_k^n \\
    \boldsymbol{v}_k^n \\
    \boldsymbol{C}_{b, k}^n
    \end{array}\right]=\left[\begin{array}{c}
    \boldsymbol{r}_{k-1}^n+\boldsymbol{v}_k^n \Delta t \\
    \boldsymbol{v}_{k-1}^n+\left[\boldsymbol{C}_{b, k}^n\left(\tilde{\boldsymbol{f}}_k^b-\boldsymbol{b}_a\right)+\boldsymbol{g}^n\right] \Delta t \\
    \boldsymbol{C}_{b, k-1}^n+\boldsymbol{C}_{b, k-1}^n\left[\left(\tilde{\boldsymbol{\omega}}_k^b-\boldsymbol{b}_g\right) \times\right] \Delta t
    \end{array}\right]
\end{equation}
where $\boldsymbol{r}^n$ and $\boldsymbol{v}^n$ represent the position vector and the velocity vector in the n-frame respectively, $\boldsymbol{C}_b^n$ represents the transformation matrix from the b-frame to the n-frame, $\boldsymbol{g}^n$ represents the Earth gravity vector in the n-frame, $\tilde{\boldsymbol{f}}_k^b$ and $\tilde{\boldsymbol{\omega}}_k^b$ represent the acceleration measurement vector and angle rate measurement in the b-frame, respectively. $\boldsymbol{b}_a$ and $\boldsymbol{b}_g$ represent the bias of the accelerometer and gyroscope, respectively. $\Delta t = t_k - t_{k-1}$ represents the time interval between the $k$-th and $(k-1)$-th epoch.

\subsection{Multi State Constraint Extended Kalman Filter}
We utilize the Multi-State Constraint Extended Kalman Filter (MSC‑EKF) for data fusion. MSC‑EKF is a filtering‑based visual‑inertial odometry (VIO) method \cite{ref11} that maintains a sliding window of state vectors spanning multiple epochs, encompassing both current and historical states (e.g., position and attitude). It leverages constraint relationships between historical states and current observations during measurement updates. In the proposed MSC‑EKF, we incorporate multiple historical position states into the state vector. Consequently, we define the error state vector as follows:

\begin{small}
    \begin{equation} \label{eq24}
    \delta \boldsymbol{x}_k = \left[\begin{array}{llllllll}
    \delta \boldsymbol{r}_k^n & \delta \boldsymbol{v}_k^n & \boldsymbol{\phi}_k & \delta \boldsymbol{b}_{g, k} & \delta \boldsymbol{b}_{a, k} & \delta \boldsymbol{r}_{k-m}^n & \cdots & \delta \boldsymbol{r}_{k-1}^n
    \end{array}\right]^{\top}
    \end{equation}
\end{small}

where $\delta \boldsymbol{r}^n$,$\delta \boldsymbol{v}^n$ and $\boldsymbol{\phi}$ represent the error vectors of position, velocity, and attitude in the n-frame, respectively. $\delta\boldsymbol{b}_{g}$ and $\delta\boldsymbol{b}_{a}$ represent the error vectors for gyroscope and accelerometer biases, respectively. $k$ denotes the epoch index, $m$ specifies the number of historical states retained in the sliding window, and $\delta \boldsymbol{r}_{k-m}^n$ and $\delta \boldsymbol{r}_{k-1}^n$ represent the position error vectors of the historical states. The discretized and linearized system error model can be formulated as follows:
\begin{equation} \label{eq25}
    \left\{\begin{aligned}
    \delta \boldsymbol{x}_{k \mid k-1} & =\boldsymbol{\Phi}_{k-1} \delta \boldsymbol{x}_{k-1 \mid k-1}+\boldsymbol{w}_k \\
    \delta \boldsymbol{z}_k & =\boldsymbol{H}_k \delta \boldsymbol{x}_{k \mid k-1}+\boldsymbol{n}_k
    \end{aligned}\right.
\end{equation}
where the subscripts $k-1$ and $k$ represent the epoch, $\delta \boldsymbol{x}_{k-1 \mid k-1}$ and $\delta \boldsymbol{x}_{k \mid k-1}$ represent the previous and predicted error state vectors, respectively. $\delta \boldsymbol{z}_k$ represents the measurement misclosure vector, $\boldsymbol{H}_k$ is the observation matrix, $\boldsymbol{w}_k$ and $\boldsymbol{n}_k$ are the process noise and measurement noise, respectively. The state transition matrix $\boldsymbol{\Phi}_{k}$ is expressed as follows:

\begin{small}
    \begin{equation} \label{eq26}
        \boldsymbol{\Phi}_{k, 15 \times 15} =\left[\begin{array}{ccccc}
        \boldsymbol{I}_{33} & \boldsymbol{I}_{33} \Delta t & \boldsymbol{0}_{33} & \boldsymbol{0}_{33} & \boldsymbol{0}_{33} \\
        \boldsymbol{0}_{33} & \boldsymbol{I}_{33} & \left(\boldsymbol{f}_k^n \times\right) \Delta t & \boldsymbol{0}_{33} & \boldsymbol{C}_{b, k}^n \Delta t \\
        \boldsymbol{0}_{33} & \boldsymbol{0}_{33} & \boldsymbol{I}_{33} & -\boldsymbol{C}_{b, k}^n \Delta t & \boldsymbol{0}_{33} \\
        \boldsymbol{0}_{33} & \boldsymbol{0}_{33} & \boldsymbol{0}_{33} & \boldsymbol{I}_{33} & \boldsymbol{0}_{33} \\
        \boldsymbol{0}_{33} & \boldsymbol{0}_{33} & \boldsymbol{0}_{33} & \boldsymbol{0}_{33} & \boldsymbol{I}_{33}
        \end{array}\right]
    \end{equation}
\end{small}

\begin{equation} \label{eq27}
    \boldsymbol{\Phi}_{k} = \left[\begin{array}{cc}
        \boldsymbol{\Phi}_{k, 15 \times 15} & \boldsymbol{0}_{15 \times 3m} \\
        \boldsymbol{0}_{3m \times 15} & \boldsymbol{I}_{3m \times 3m}
    \end{array}\right 
    ]
\end{equation}

In the MSC-EKF framework, cloned states(augmented historical states) do not require updates during the one-step prediction phase. After state prediction, the covariance propagation formula is expressed as:
\begin{equation} \label{eq28}
    \boldsymbol{P}_{k,k-1} = \boldsymbol{\Phi}_{k-1} \boldsymbol{P}_{k-1} \boldsymbol{\Phi}_{k-1}^\top + \boldsymbol{Q}_k
\end{equation}
where $\boldsymbol{P}_{k-1}$ denotes the initial state covariance matrix, $\boldsymbol{Q}_k$ denotes the process noise covariance matrix for state prediction. 
When valid observations are acquired, the following formula can be applied to update the state vector and its associated covariance matrix:
\begin{equation} \label{eq29}
    \delta\boldsymbol{x}_k = \delta\boldsymbol{x}_{k,k-1} + \boldsymbol{K}_k\left(\delta \boldsymbol{z}_k - \boldsymbol{H}_k \boldsymbol{x}_{k,k-1}\right)
\end{equation}
\begin{equation} \label{eq30}
    \boldsymbol{P}_k = \left(\boldsymbol{I} - \boldsymbol{K}_k \boldsymbol{H}_k \right) \boldsymbol{P}_{k,k-1} {\left(\boldsymbol{I} - \boldsymbol{K}_k \boldsymbol{H}_k \right)}^\top + \boldsymbol{K}_k \boldsymbol{R}_k \boldsymbol{K}_k^{\top}
\end{equation}
\begin{equation} \label{eq31}
    \boldsymbol{K}_k = \boldsymbol{P}_{k,k-1} \boldsymbol{H}_k^{\top} \left(\boldsymbol{H}_k \boldsymbol{P}_{k,k-1} \boldsymbol{H}_k^{\top} + \boldsymbol{R}_k \right)^{-1}
\end{equation}

It is noteworthy that after the MSC-EKF completes the state vector update, a block permutation of the current covariance matrix $\boldsymbol{P}_k$ must be performed. This step aims to manage the $m$ historical cloned states in the sliding window and ensure structural consistency between the covariance matrix and the state vector. Taking $m=2$ as an example, the specific cloning strategy for $\boldsymbol{P}_k$ is as follows:
\begin{equation} \label{eq32}
    \boldsymbol{P}_k^{new} = \boldsymbol{H}_{clone} \boldsymbol{P}_k \boldsymbol{H}_{clone}^{\top}
\end{equation}
\begin{equation} \label{eq33}
    \boldsymbol{H}_{clone}=\left[\begin{array}{llll}
    \boldsymbol{I}_{15 \times 15} & & \mathbf{0}_{3 \times 3} & \mathbf{0}_{3 \times 3} \\
    & & \mathbf{0}_{12 \times 3} & \mathbf{0}_{12 \times 3} \\
    \mathbf{0}_{3 \times 3} & \mathbf{0}_{3 \times 12} & \mathbf{0}_{3 \times 3} & \boldsymbol{I}_{3 \times 3} \\
    \boldsymbol{I}_{3 \times 3} & \mathbf{0}_{3 \times 12} & \mathbf{0}_{3 \times 3} & \mathbf{0}_{3 \times 3}
    \end{array}\right]
\end{equation}

\subsection{Observation model}
As derived in Section II, this paper has established the local magnetic field model and its predictive formulation across adjacent epochs. After performing an error perturbation analysis on Equation \eqref{eq22}, the predicted magnetic field model can be expressed as:
\begin{equation} \label{eq34}
    \begin{aligned}
        \hat{\boldsymbol{M}}_j^{b_j} &= \boldsymbol{M}_j^{b_j} + \boldsymbol{C}_i^j \boldsymbol{B}(\boldsymbol{\theta})\delta\boldsymbol{l}_j^{b_i} + [\boldsymbol{M}_j^{b_i}\times] \delta\boldsymbol{b}_g {dt} \\
        &= \boldsymbol{M}_j^{b_j} + \boldsymbol{C}_i^j \boldsymbol{B}(\boldsymbol{\theta}) \boldsymbol{C}_n^{b_i} \delta\boldsymbol{v}_j^n {dt} \\ 
        &+ \left([\boldsymbol{M}_j^{b_i}\times] - \boldsymbol{C}_i^j \boldsymbol{B}(\boldsymbol{\theta}) [\boldsymbol{l}_j^{b_j}\times]\right) \delta\boldsymbol{b}_g {dt} \\
        &- \boldsymbol{C}_i^j \boldsymbol{B}(\boldsymbol{\theta}) [\Delta\boldsymbol{r}^{b_i}\times] \boldsymbol{\phi} 
    \end{aligned}
\end{equation}
where $\Delta t$ denotes the time difference between adjacent epochs. Taking the first-order magnetic field model as an example, $\boldsymbol{B}(\boldsymbol{\theta})$ is formulated as:
\begin{equation} \label{eq35}
    \boldsymbol{B}(\boldsymbol{\theta}) = \left[\begin{array}{ccc}
        2\theta_4 & \theta_6 & \theta_7 \\
        \theta_6 & 2\theta_5 & \theta_8 \\
        \theta_7 & \theta_8 & -2(\theta_4 + \theta_5)
    \end{array}\right]
\end{equation}

For the magnetic field model from the previous epoch, the corresponding observations can be obtained via magnetometer array measurements as follows:
\begin{equation} \label{eq36}
    \tilde{\boldsymbol{M}}_j^{b_j} = \boldsymbol{M}_j^{b_j} + \boldsymbol{n}_v
\end{equation}
where $\boldsymbol{n}_v$ denotes the observation noise. By combining Equations \eqref{eq34} and \eqref{eq36}, the following observation equation can be derived:
\begin{equation} \label{eq37}
    \begin{aligned}
        \delta \boldsymbol{z}_{mag} &= \hat{\boldsymbol{M}}_j^{b_j} - \tilde{\boldsymbol{M}}_j^{b_j} \\
        &= \boldsymbol{C}_i^j \boldsymbol{B}(\boldsymbol{\theta}) \boldsymbol{C}_n^{b_i} \delta\boldsymbol{v}_j^n {dt} \\
        &+ \left([\boldsymbol{M}_j^{b_i}\times] - \boldsymbol{C}_i^j \boldsymbol{B}(\boldsymbol{\theta}) [\boldsymbol{l}_j^{b_j}\times]\right) \delta\boldsymbol{b}_g {dt} \\ 
        &- \boldsymbol{C}_i^j \boldsymbol{B}(\boldsymbol{\theta}) [\Delta\boldsymbol{r}^{b_i}\times] \boldsymbol{\phi} + \boldsymbol{n}_v
    \end{aligned}
\end{equation}

As derived in the preceding sections, the observation equation \eqref{eq37} utilizing magnetometer array measurements for the filter's measurement update has been established. Therefore, the design matrix can be formulated as:
\begin{equation} \label{eq38}
    \boldsymbol{H}_{mag} = \left[\begin{array}{cccccccc}
        \boldsymbol{0}_{33} & \boldsymbol{H}_{12} & \boldsymbol{H}_{13} & \boldsymbol{H}_{14} & \boldsymbol{0}_{33} & \boldsymbol{0}_{33} & ... & \boldsymbol{0}_{33}
    \end{array}\right]
\end{equation}
The submatrix in $\boldsymbol{H}_{mag}$ is defined as:
\begin{equation} \label{eq39}
    \left\{\begin{aligned}
         \boldsymbol{H}_{12} &= \boldsymbol{C}_i^j \boldsymbol{B}(\boldsymbol{\theta}) \boldsymbol{C}_n^{b_i} {dt} \\
         \boldsymbol{H}_{13} &= - \boldsymbol{C}_i^j \boldsymbol{B}(\boldsymbol{\theta}) [\Delta\boldsymbol{r}^{b_i}\times] \\
         \boldsymbol{H}_{14} &= - \boldsymbol{C}_i^j \boldsymbol{B}(\boldsymbol{\theta}) [\boldsymbol{l}_j^{b_j}\times] {dt} 
    \end{aligned}\right.
\end{equation}
Since $\delta\boldsymbol{z}_{mag}$ incorporates both measurement noise and estimation errors in the magnetic field model parameters, the observation covariance matrix should be composed of two components:
\begin{equation} \label{eq40}
    \boldsymbol{R}_{meas} = \left(\boldsymbol{C}_i^j \boldsymbol{\Phi}(\boldsymbol{l}_j^{b_i})\right) \boldsymbol{Cov}_{\hat{\boldsymbol{\theta}}}\left(\boldsymbol{C}_i^j \boldsymbol{\Phi}(\boldsymbol{l}_j^{b_i})\right)^{\top} + \text{diag}(\sigma_{mag}^2)
\end{equation}
Here $\sigma_{mag}^2$ denotes the standard deviation of the magnetometer array measurement noise.
\par
Previous studies \cite{ref12} have shown a mismatch between estimated state covariance and actual system uncertainty in EKF‑based odometry‑aided INS implementations. Likewise, in magnetometer array–aided INS scenarios, the heading angle becomes unobservable due to the absence of absolute heading references. However, the EKF’s Jacobian linearization introduces higher‑order truncation errors, resulting in spurious observability of inherently unobservable states within the covariance matrix. This effect artificially inflates the filter’s confidence in heading error estimates, potentially causing filter divergence. 

To address heading unobservability, we introduce a magnetic vector attitude constraint in the navigation frame. This constraint uses differences between consecutive n‑frame magnetic field measurements from a single magnetometer, enabling measurement updates driven by attitude variations. Assuming (1) the n‑frame magnetic vector remains constant and (2) the attitude error is unchanged between adjacent epochs, the observed n‑frame magnetic vector can be expressed as follows:

\begin{equation} \label{eq41}
\begin{aligned}
    \hat{\boldsymbol{M}}_i^n &= \hat{\boldsymbol{C}}_b^n \boldsymbol{M}_i^b = [\boldsymbol{I}-\boldsymbol{\phi} \times] \boldsymbol{C}_b^n \boldsymbol{M}_i^b \\
    \hat{\boldsymbol{M}}_j^n &= \hat{\boldsymbol{C}}_b^n \boldsymbol{M}_j^b = [\boldsymbol{I}-\boldsymbol{\phi} \times] \boldsymbol{C}_b^n \boldsymbol{M}_j^b
\end{aligned}
\end{equation}
The Kalman filter measurement equation incorporating the magnetic vector attitude constraint is formulated as:
\begin{equation} \label{eq42}
    \delta\boldsymbol{z} = \hat{\boldsymbol{M}}_i^n - \hat{\boldsymbol{M}}_j^n = \left[  (\boldsymbol{M}_i^b - \boldsymbol{M}_j^b) \times \right] \boldsymbol{\phi}
\end{equation}

where $\boldsymbol{M}_i^b$ denotes the magnetic field vector in the body frame at the current epoch; $\boldsymbol{M}_j^b$ denotes the magnetic field vector in the b-frame at the previous epoch; $\boldsymbol{C}_b^n$ represents the rotation matrix transforming vectors from the b-frame to the navigation frame. The corresponding design matrix is expressed as:

\begin{small}
    \begin{equation} \label{eq43}
        \boldsymbol{H}_{att} = \left[\begin{array}{cccccccc}
            \boldsymbol{0}_{33} & \boldsymbol{0}_{33} & \left[  (\boldsymbol{M}_i^b - \boldsymbol{M}_j^b) \times \right] & \boldsymbol{0}_{33} & ... & \boldsymbol{0}_{33}
        \end{array}\right]
    \end{equation}    
\end{small}

\section{Experimental Results}
\subsection{Results on OA dataset}
To validate the performance of the proposed MSCEKF-MIO algorithm, this section conducts a comparative analysis with the state-of-the-art open-source magnetic-odometry algorithm MAINS, leveraging the open dataset from \cite{ref2}. In this comparison, MAINS employs the rectangular sensor configuration with 30 magnetometers. Fig.3 present the trajectory estimation results of both algorithms on the public dataset. Tab.\ref{tab:table1} summarizes the Root Mean Square (RMS) horizontal position and velocity errors for the two algorithms across the dataset. 

\begin{figure*}[!htbp]
    \centering
    \subfloat[\sffamily\footnotesize LP-1]{\includegraphics[width=0.3\linewidth, trim=0 80 0 160]{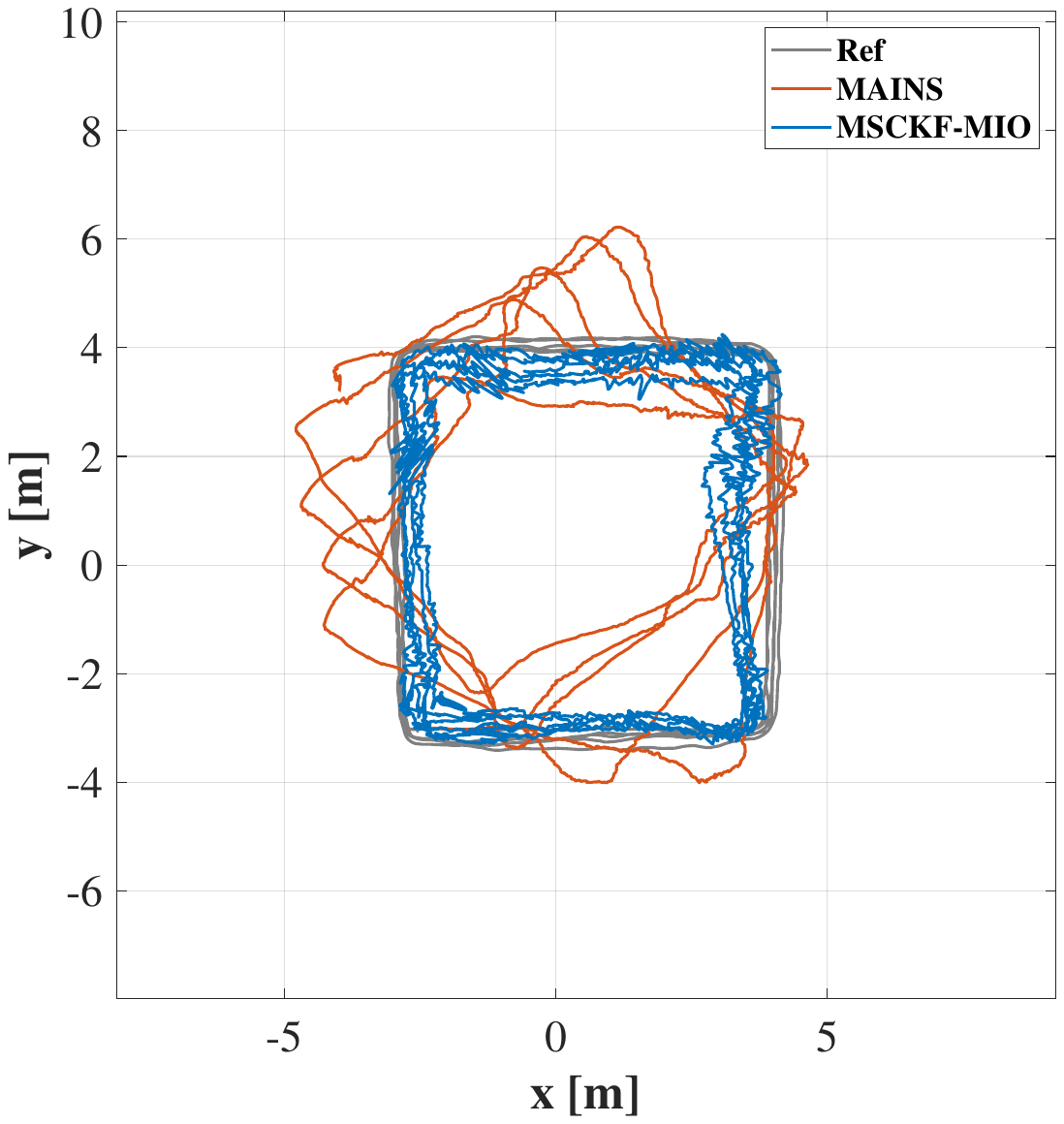}}
    \hfill
    \subfloat[\sffamily\footnotesize LP-2]{\includegraphics[width=0.3\linewidth, trim=0 80 0 160]{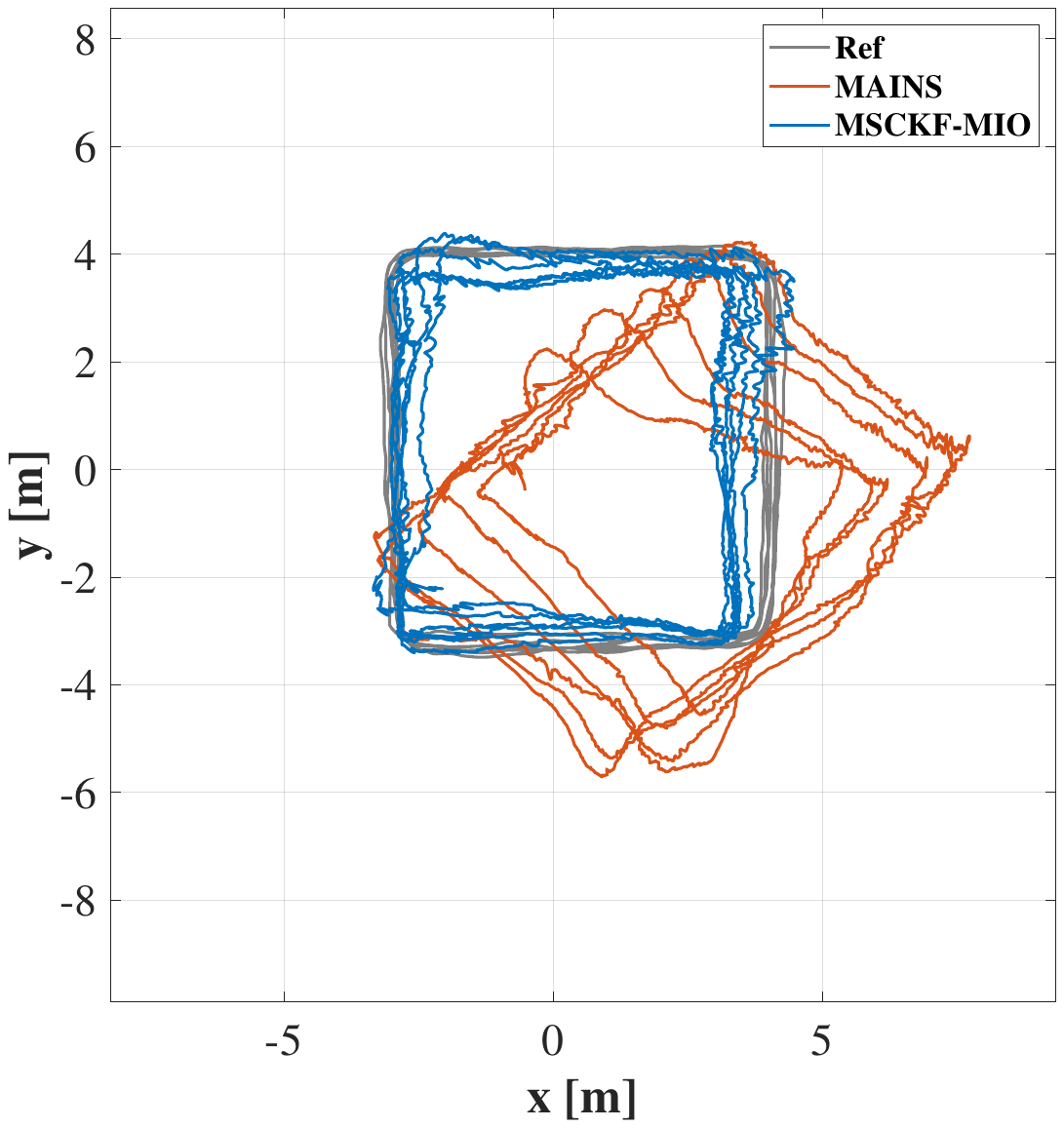}}
    \hfill
    \subfloat[\sffamily\footnotesize LP-3]{\includegraphics[width=0.3\linewidth, trim=0 80 0 160]{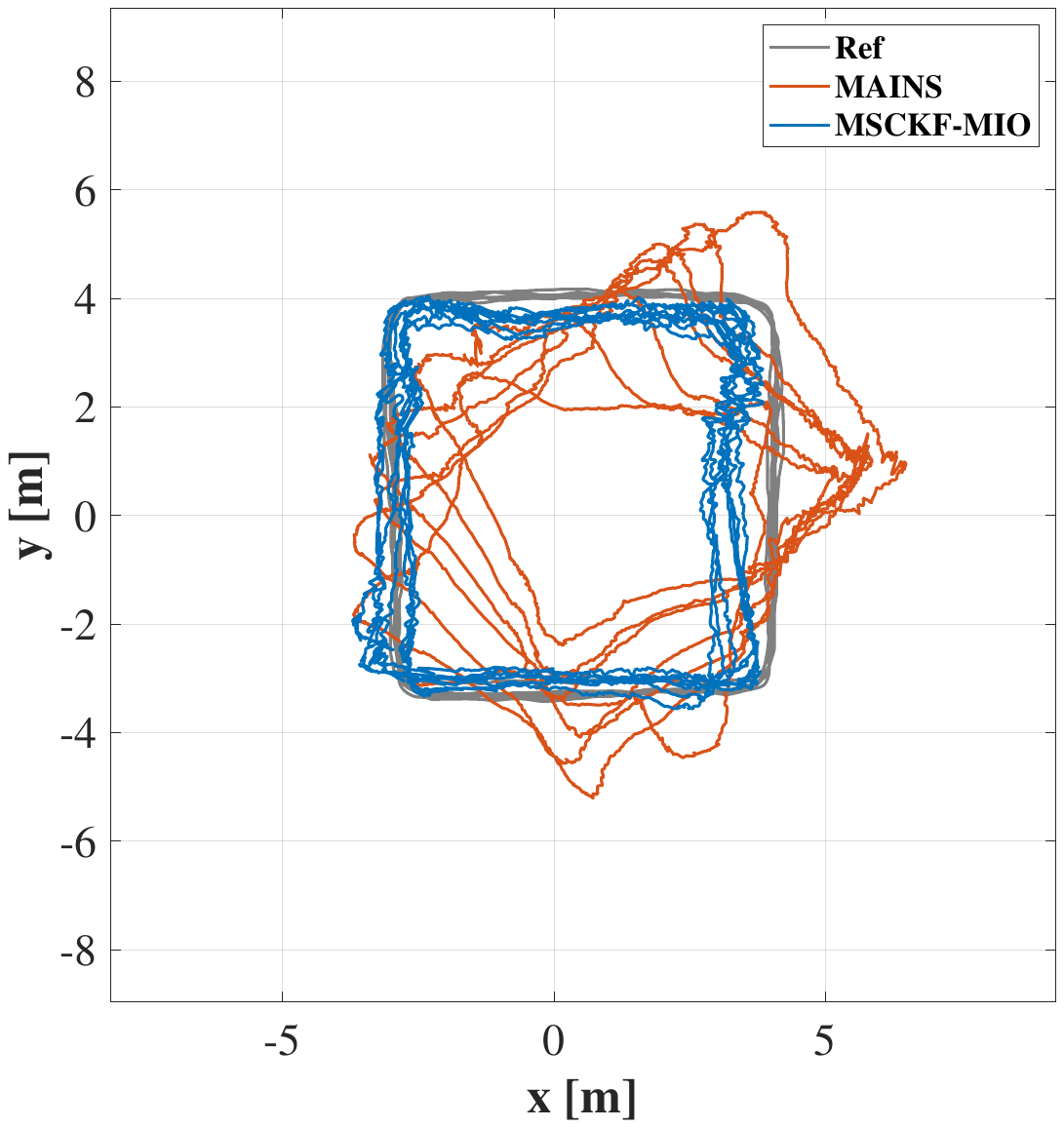}}
    \caption{Trajectory estimation results of OA dataset}
    \label{fig:oa_results}
\end{figure*}

Fig.3 demonstrates the trajectory estimation comparison under a scenario with an average altitude of 0.50 m. The comparison reveals that MAINS exhibits significant cumulative position errors during multi-loop motion, accompanied by rapid divergence in heading angle estimates. In contrast, the proposed MSCEKF-MIO algorithm effectively suppresses position error accumulation through its enhanced magnetic field model and multi-state constraint mechanism, maintaining tight adherence to the reference trajectory within bounded deviations. 

According to the statistical results, the proposed MSCEKF-MIO achieves an RMS horizontal position error of approximately 0.50 m on the dataset with an average altitude of 0.50 m, representing an 85\% improvement compared to MAINS (3.44 m). Notably, in velocity estimation, MSCEKF-MIO reduces the RMS error by 40\% relative to MAINS, demonstrating that the enhanced magnetic field model and multi-state constraints effectively improve the accuracy of velocity estimation. Comparisons of localization results and statistical metrics for the remaining public datasets, along with detailed numerical values, are provided in Tab.\ref{tab:table6} and Fig.\ref{fig:other_results}. These results confirm that the proposed MSCEKF-MIO algorithm achieves high-precision localization on open datasets without relying on external positional references (e.g., the 60-second initial positional assistance required by MAINS).

Fig.\ref{fig:other_results} shows both algorithms’ localization results diverging significantly in position and heading. This behavior arises because the dataset’s elevation remains between 0.70 m and 0.80 m. We hypothesize that, at this elevation, the magnetic field gradient attenuates substantially, causing the gradient variations detected by the magnetometer array to flatten. As a result, this flattening introduces errors into the perturbation model linking magnetic gradient variations to pose changes, ultimately causing localization divergence.

\begin{table}[!t]
\caption{Error Statistics of Public Dataset-1\label{tab:table1}}
\centering
\begin{threeparttable}
\begin{tabular}{cccc}
\hline
\hline
Data sequence & \textbf{LP-1} & \textbf{LP-2} & \textbf{LP-3}  \\
\hline
Trajectory length (m) & 114.14 & 139.87 & 192.93 \\
\hline
Trajectory duration (s) & 212 & 226 & 332 \\
\hline
Average height (m) & 0.49 & 0.52 & 0.53 \\
\hline
\rowcolor{gray!20}  \multicolumn{4}{c}{\textbf{MAINS}} \\
\hline
RMS (m) & 3.15 & 3.61 & 3.56 \\
\hline
CDF68 (m) & 3.36 & 4.11 & 3.75 \\
\hline
RMS Speed (m/s) & 0.09 & 0.10 & 0.12\\
\hline
\rowcolor{gray!20}  \multicolumn{4}{c}{\textbf{MSCEKF-MIO}} \\
\hline
RMS (m) & 0.49 & 0.58 & 0.58 \\
\hline
CDF68 (m) & 0.53 & 0.66 & 0.61 \\
\hline
RMS Speed (m/s) & 0.06 & 0.07 & 0.07\\
\hline
\end{tabular}
    \begin{tablenotes}
        \footnotesize
        \item[1] LP: low height and parallel  NP: normal height and parallel  NT: normal height and tilted.
    \end{tablenotes}
\end{threeparttable}
\end{table}

\subsection{Results on Our dataset}
\begin{figure}
    \centering
    \includegraphics[scale=0.18]{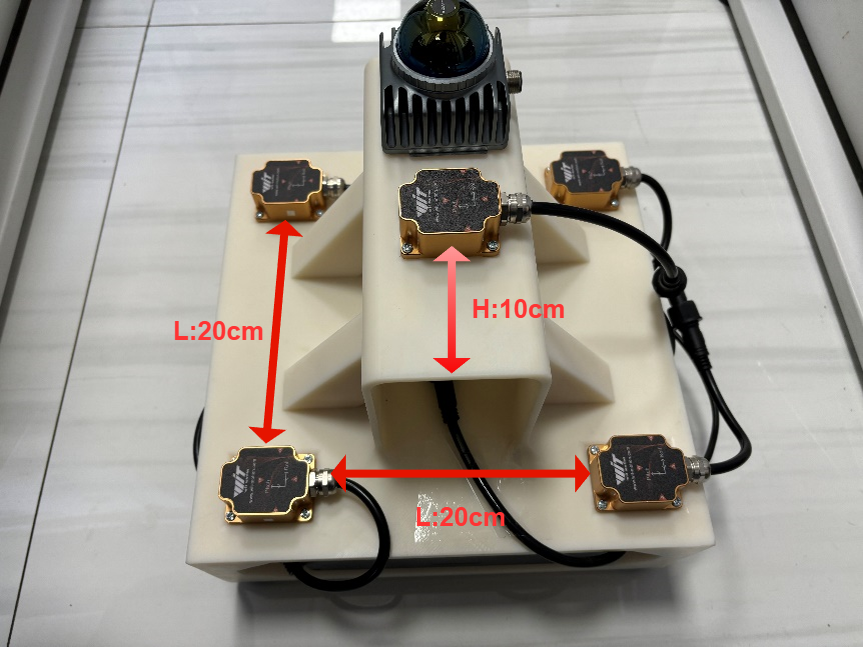}
    \caption{Sensor array platform}
    \label{fig:platform}
\end{figure}

To further evaluate the performance of the proposed algorithm, a custom-developed sensor array platform (as illustrated in Fig.\ref{fig:platform}) was constructed, and extensive experimental data was collected across multiple indoor scenarios. Scenario a: office building lobby, Scenario b: underground parking garage, Scenario c: indoor corridor.

\begin{figure}[!htbp]
    \centering
    \subfloat[\sffamily\footnotesize Scenario-a]{\includegraphics[width=0.3\linewidth]{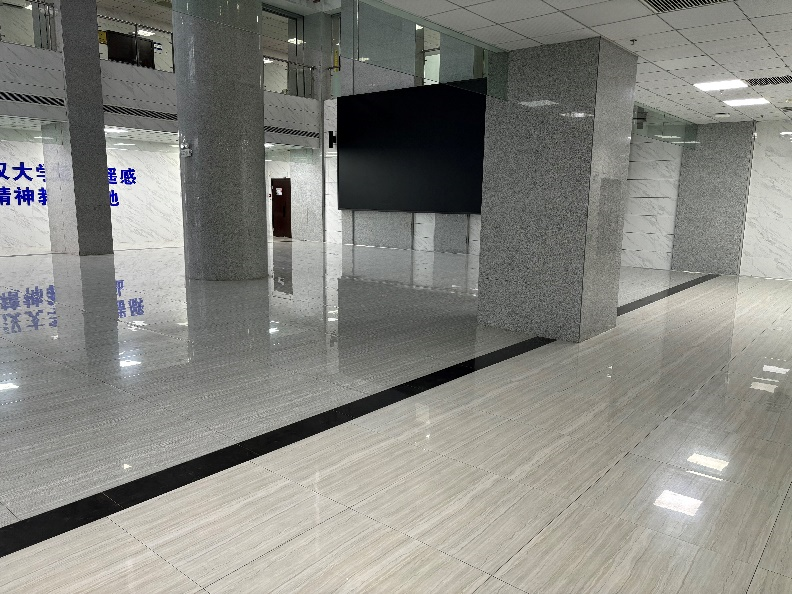}}
    \label{fig:sc_a}
    \hfill
    \subfloat[\sffamily\footnotesize Scenario-b]{\includegraphics[width=0.3\linewidth]{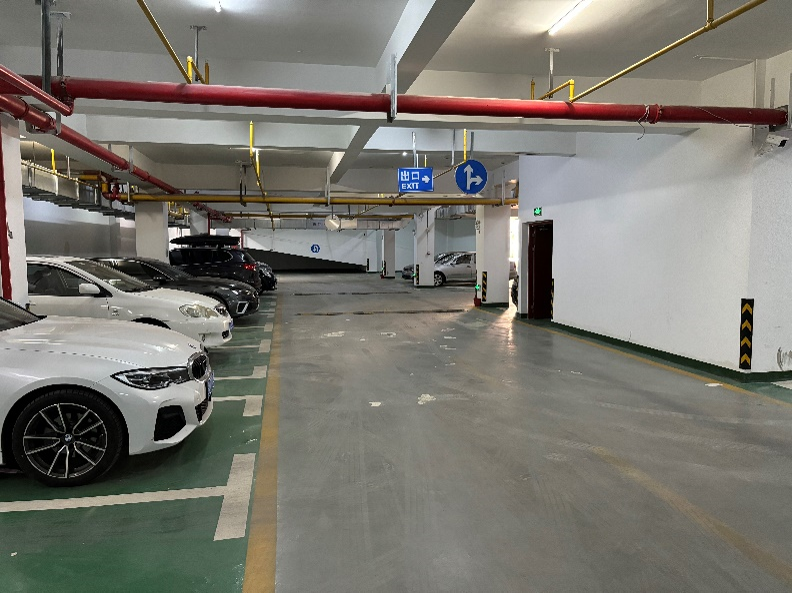}}
    \label{fig:sc_b}
    \hfill
    \subfloat[\sffamily\footnotesize Scenario-c]{\includegraphics[width=0.3\linewidth]{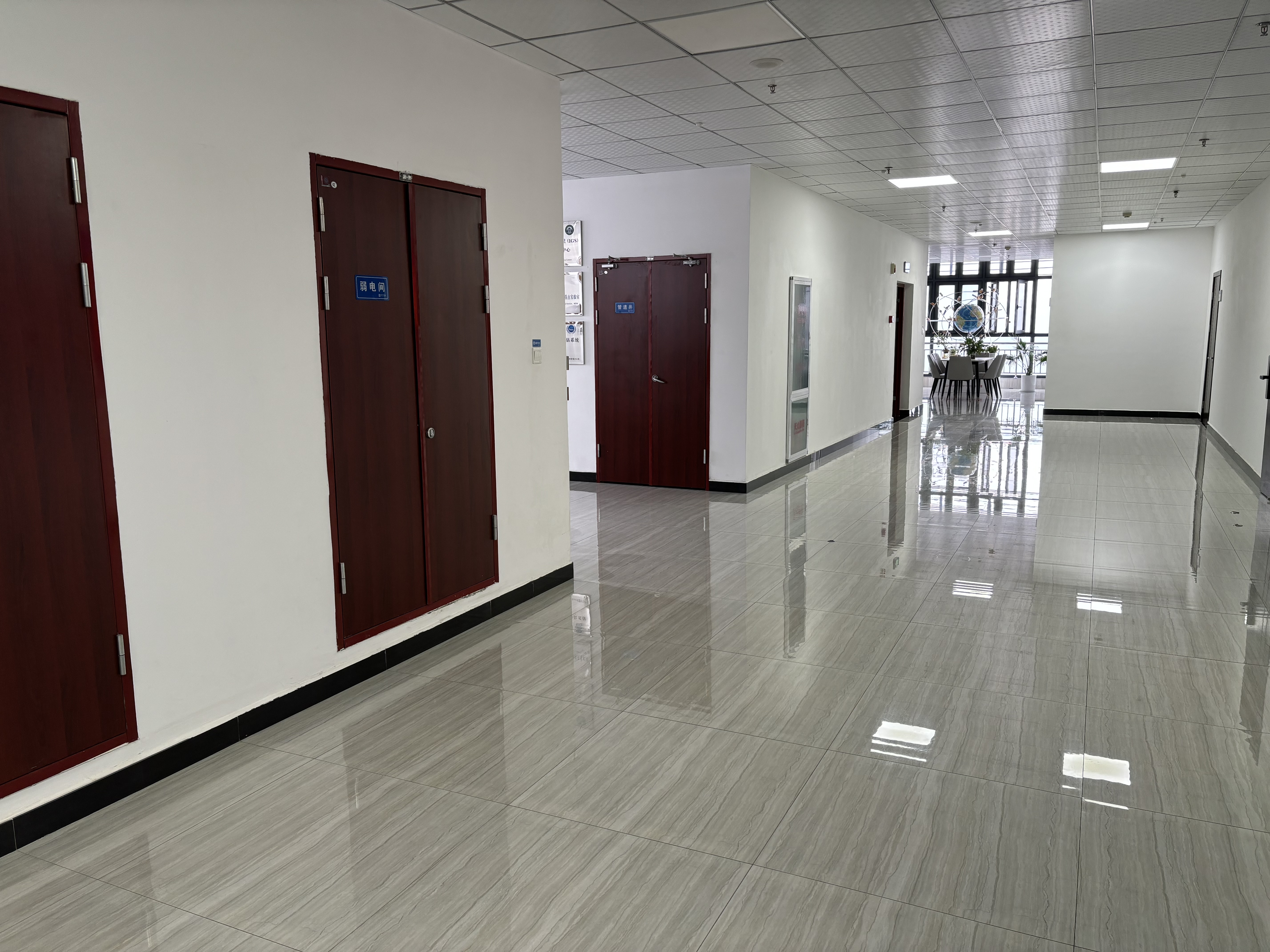}}
    \label{fig:sc_c}
    \caption{Data acquisition environment}
    \label{fig:oa_results}
\end{figure}
The sensor array platform is equipped with five 9-axis sensor modules (HWT-9073) and one LiDAR (Livox Mid-360). Each HWT-9073 module integrates a 3-axis gyroscope, 3-axis accelerometer, and 3-axis magnetometer. The Livox Mid-360 LiDAR captures point cloud data and is paired with the Fast-LIO2 algorithm \cite{ref13} to compute reference poses for indoor pedestrian tracking, achieving decimeter-level accuracy.

Magnetometers are prone to interference from onboard or environmental magnetic disturbances, particularly soft-iron effects, which can cause misalignment between the magnetometer and inertial sensor frames.  Therefore, magnetometer calibration and alignment with inertial sensors are necessary before using the magnetometer array for data collection. In the data preprocessing stage of this experiment, a Kalman filter-based magnetometer calibration and alignment algorithm from \cite{ref14} was applied to ensure measurement accuracy. For the experiments, a pedestrian acted as the carrier of the magnetic-inertial odometry system. During data collection, the pedestrian held the platform and maintained it at a fixed height above the ground. A total of 10 datasets were collected, with the characteristics of each dataset detailed in Tab.\ref{tab:table2}.

\begin{figure*}[b]
    \centering
    \subfloat[\sffamily\footnotesize aM-1]{\includegraphics[width=0.3\linewidth, trim=0 80 0 160]{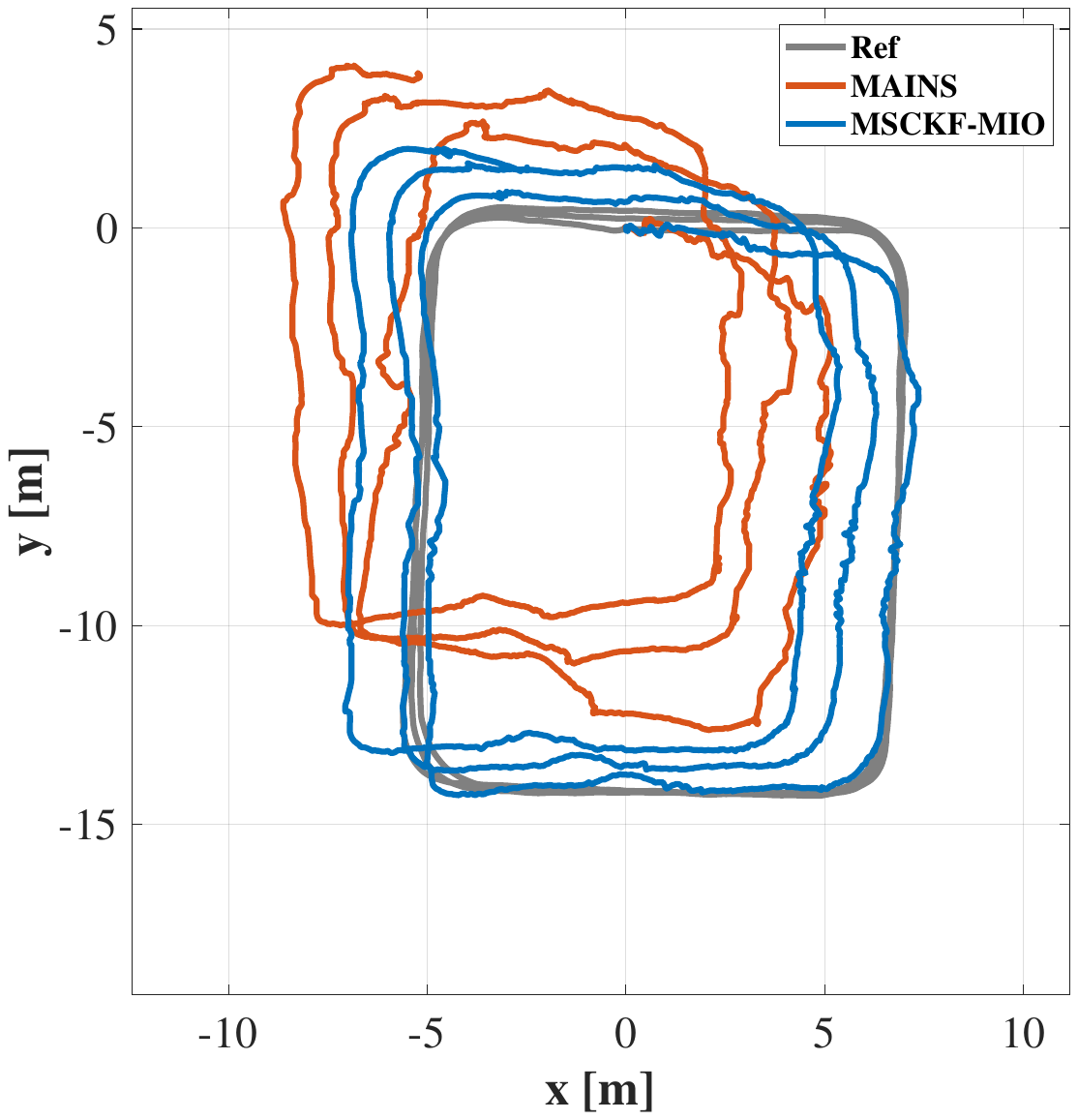}}
    \label{fig:am1}
    \hfill
    \subfloat[\sffamily\footnotesize bM-1]{\includegraphics[width=0.3\linewidth, trim=0 80 0 160]{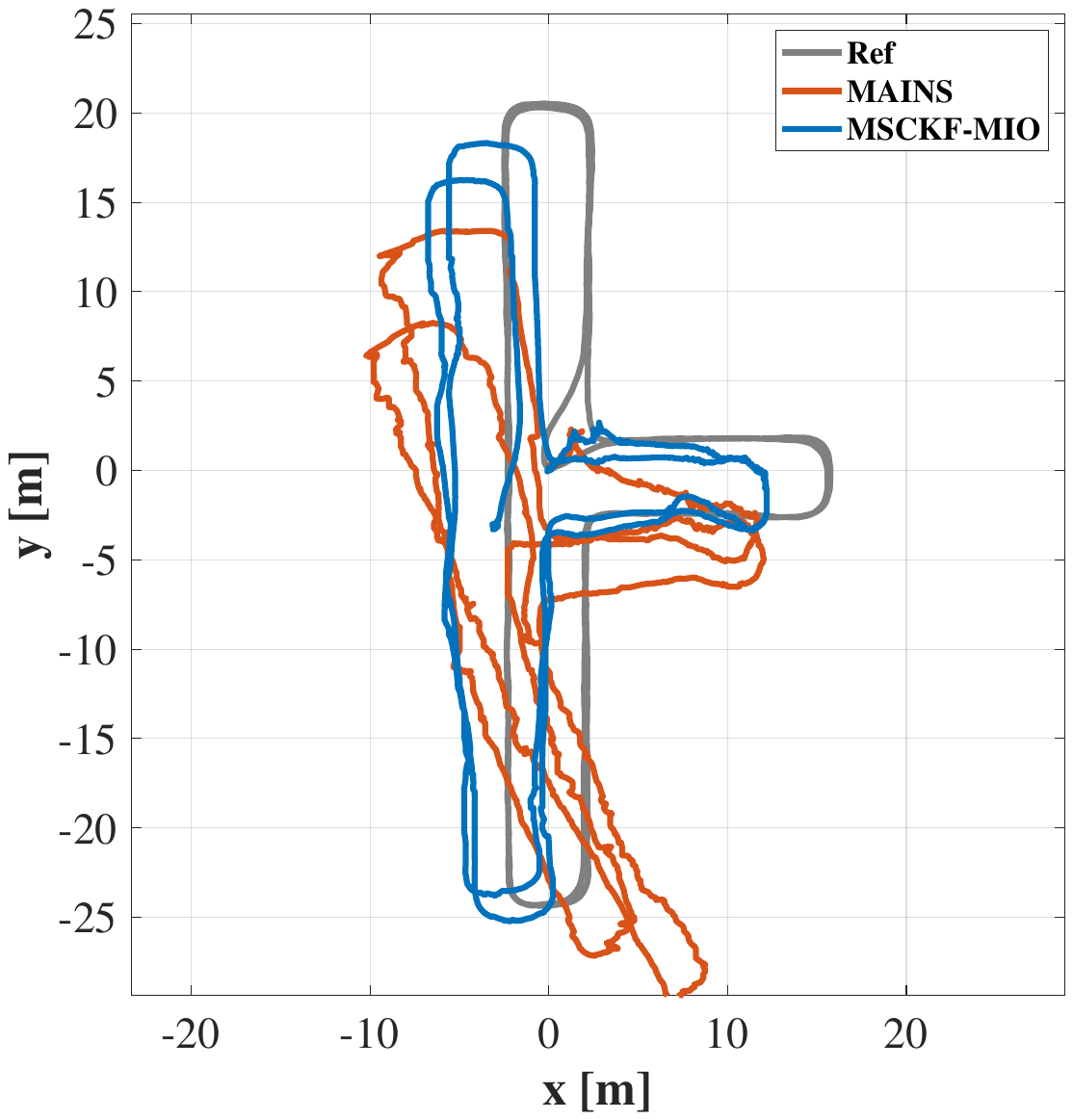}}
    \label{fig:bm1}
    \hfill
    \subfloat[\sffamily\footnotesize cM-1]{\includegraphics[width=0.3\linewidth, trim=0 80 0 160]{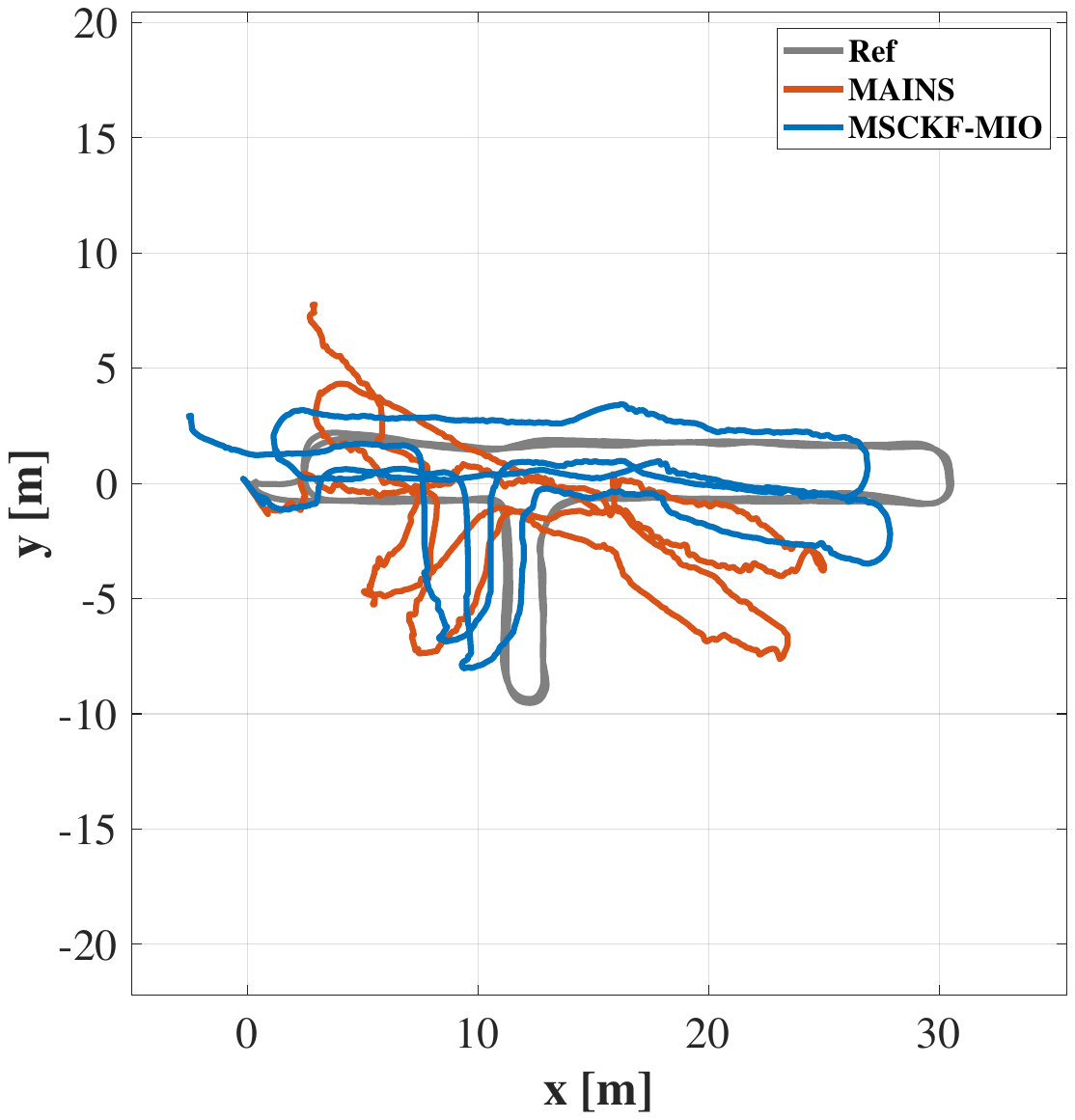}}
    \label{fig:cm1}
    \caption{Trajectory estimation results of our dataset}
    \label{fig:our_results}
\end{figure*}

\begin{table}[!t]
\caption{INFORMATION ABOUT THE DATASETS\label{tab:table2}}
\centering
\begin{threeparttable}
\begin{tabular}{cccc}
\hline
\textbf{Data sequence} & \textbf{Average height(m)} & \textbf{Length(m)} & \textbf{Duration(s)}  \\
\hline
aL-1 & 0.41 & 146.32 & 156 \\
\hline
aL-2  &	0.40  &	143.60  & 	145 \\
\hline
aM-1  &	0.54  &	149.79 & 177  \\
\hline
aM-2  &	0.68  &	148.49 & 155 \\
\hline
aN-1  &	0.84  &	137.72 & 132 \\
\hline
aN-2  &	0.80  &	138.14 & 139 \\
\hline
bM-1  &	0.68  &	247.17 & 254 \\
\hline
bM-2  &	0.63  &	248.49 & 303 \\
\hline
cM-1  &	0.54  &	155.75 & 175 \\
\hline
cM-2  &	0.50  &	154.77 & 173 \\
\hline
\end{tabular}
\end{threeparttable}
        \begin{tablenotes}
        \footnotesize
        \item[1] L: low height  M: medium height  N: normal height
    \end{tablenotes}
\end{table}

All datasets are processed using two algorithms: MAINS proposed in \cite{ref2} and the MSCEKF-MIO algorithm presented in this work. Fig.\ref{fig:our_results} illustrate the horizontal positioning results of both algorithms across three different experimental scenarios. Tab.\ref{tab:table3} provides the statistical metrics of positioning results for two datasets per scene in the three scenarios. 

\begin{table}[!htbp]
\caption{The horizontal position and velocity error statistics of our dataset\label{tab:table3}}
\centering
\begin{tabular}{ccccccc}
\hline
\hline
Data sequence & \textbf{aM-1} & \textbf{aM-2} & \textbf{bM-1} & \textbf{bM-2} & \textbf{cM-1} & \textbf{cM-2} \\
\hline
\rowcolor{gray!20}  \multicolumn{7}{c}{\textbf{MAINS}} \\
\hline
RMS (m) & 6.09 & 9.04 & 7.61 & 12.58 & 4.66 & 3.11\\
\hline
CDF68 (m) & 6.80 & 9.79 & 7.99 & 12.94 & 4.81 & 3.39\\
\hline
RMS Speed (m/s) & 0.21 & 0.21 & 0.19 & 0.20 & 0.23 & 0.24 \\
\hline
\rowcolor{gray!20}  \multicolumn{7}{c}{\textbf{MSCEKF-MIO}} \\
\hline
RMS (m) & 1.88 & 3.13 & 3.07 & 2.77 & 2.58 & 2.48\\
\hline
CDF68 (m) & 2.23 & 4.00 & 3.33 & 3.08 & 2.98 & 3.07\\
\hline
RMS Speed (m/s) & 0.12 & 0.15 & 0.13 & 0.11 & 0.13 & 0.14\\
\hline
\end{tabular}
\end{table}

The experimental results demonstrate that the proposed MSCEKF-MIO algorithm exhibits significant advantages in horizontal positioning accuracy compared to the MAINS algorithm. As shown in Fig.\ref{fig:our_results}, under closed-loop multi-lap path testing scenarios, the MAINS algorithm displays noticeable trajectory drift (RMS error $>$4 m), whereas the MSCEKF-MIO maintains close alignment with the reference trajectory (RMS error $<$2.5 m). Quantitative comparisons in Table 3 further validate this conclusion: on datasets with trajectory lengths of 150–250 m, the 68\% cumulative distribution function (CDF68) value of horizontal position errors for MSCEKF-MIO averages approximately 2.4 m, representing a reduction of about 60\% compared to MAINS (6.3 m).

Furthermore, the trajectory estimation results of the MAINS algorithm exhibit high sensitivity to the noise parameters of the magnetic field model. When the testing environment changes, the noise parameters of the magnetic field model require recalibration to match the true magnetic conditions. In contrast, the proposed MSCEKF-MIO does not incorporate the magnetic field model as a state variable. Instead, it constructs geometric constraints using pose information from multiple epochs and performs EKF updates. By eliminating excessive reliance on magnetic field modeling and reducing the degrees of freedom in state estimation, the algorithm enhances positioning accuracy and delivers more stable localization results.

\subsection{Analysis of N-frame Mag Constraint}
\begin{figure*}[!htbp]
    \centering
    \subfloat[\sffamily\footnotesize aL-1]{\includegraphics[width=0.49\linewidth, trim=80 0 80 0]{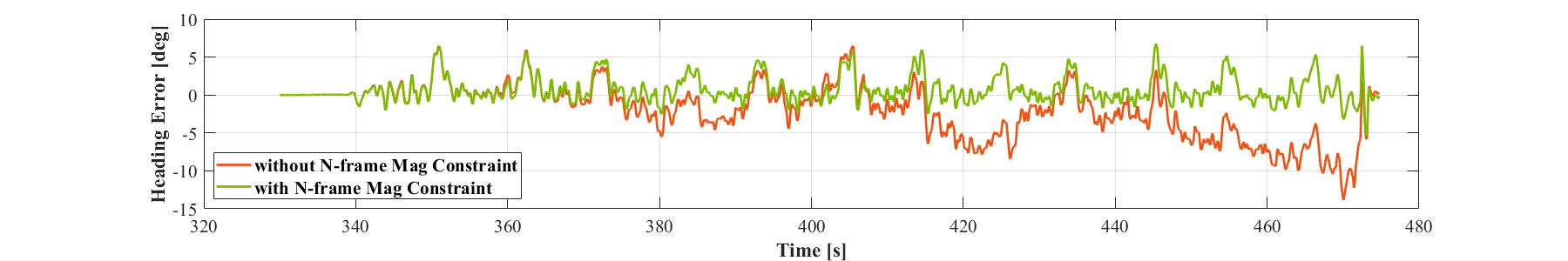}}
    \label{fig:at1}
    \hfill
    \subfloat[\sffamily\footnotesize aL-2]{\includegraphics[width=0.49\linewidth, trim=80 0 80 0]{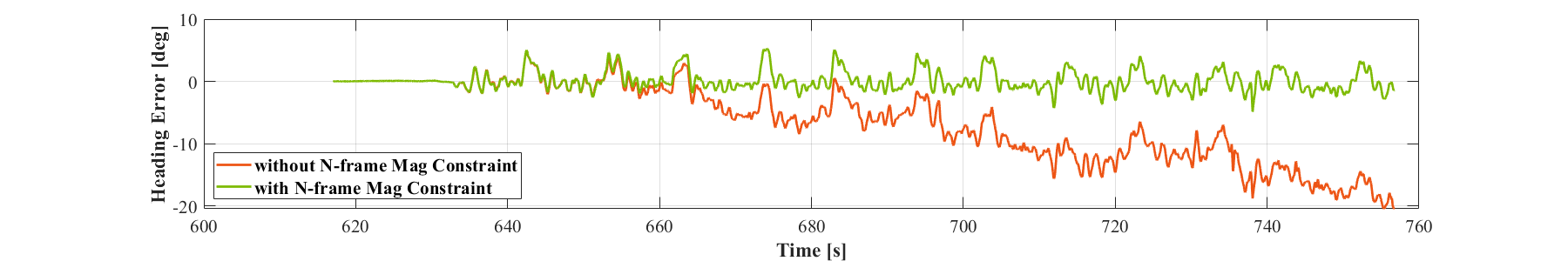}}
    \label{fig:at2}
    \subfloat[\sffamily\footnotesize aM-1]{\includegraphics[width=0.49\linewidth, trim=80 0 80 0]{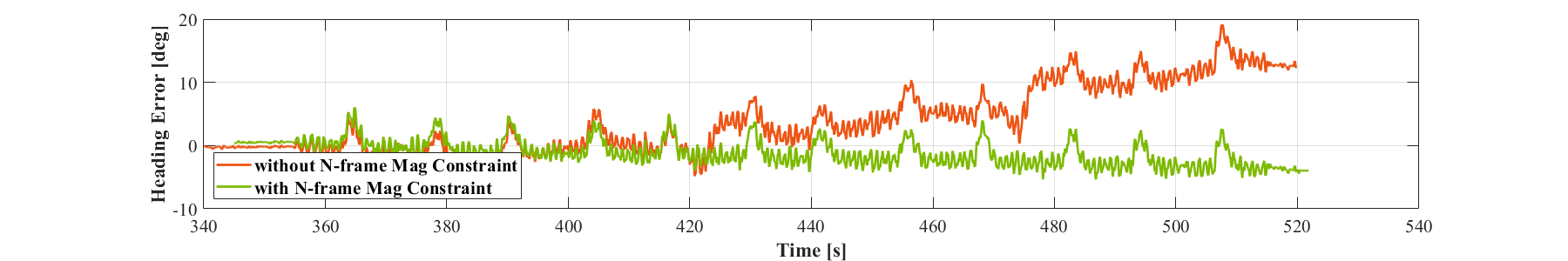}}
    \label{fig:at3}
    \hfill
    \subfloat[\sffamily\footnotesize aM-2]{\includegraphics[width=0.49\linewidth, trim=80 0 80 0]{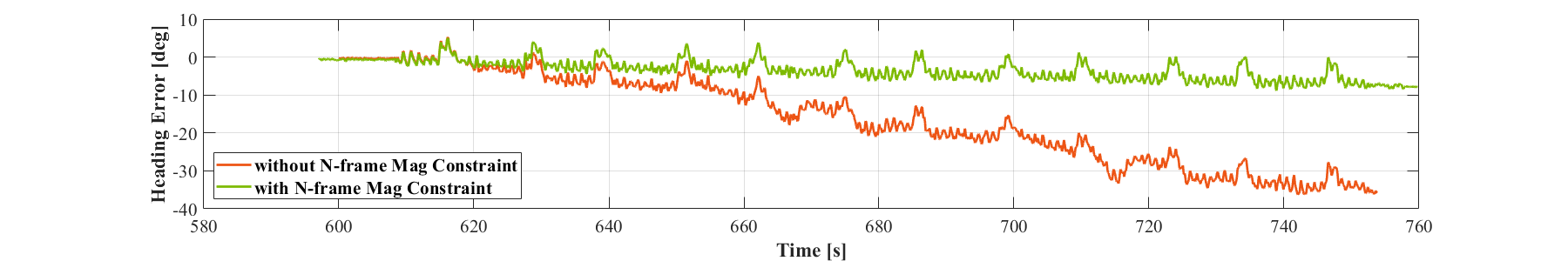}}
    \label{fig:at4}
    \caption{Comparison of Heading Angle Errors With and Without n-Frame Magnetic Vector Constraint}
    \label{fig:att_com}
\end{figure*}

To validate the optimization effect of the n-frame magnetic vector attitude constraint in the proposed algorithm on attitude estimation, this section designs a comparative experiment based on the acquired reference attitude of indoor trajectories. In the MSCEKF-MIO solution process, the n-frame magnetic vector attitude constraint is first disabled and then enabled, and the calculated heading angles are compared with the reference heading angles. The RMS values of the heading angle deviations are statistically analyzed. Fig.\ref{fig:att_com} presents the heading angle error comparisons for the four datasets (aL-1 to aM-2), and Tab.\ref{tab:table4} provides the corresponding RMS statistics of heading angle errors.

As shown in Fig.\ref{fig:att_com}, the heading angle error exhibits significant temporal divergence when the n-frame magnetic vector constraint is disabled. By the end of the trajectory, the maximum heading deviation approaches 40°. This phenomenon confirms that the heading angle in magneto-inertial odometry remains in an unobservable state in the absence of absolute heading references. In contrast, enabling the magnetic constraint substantially reduces the divergence rate of heading errors. According to the statistical metrics in Tab.\ref{tab:table4}, the RMS heading error remains consistently below 2°. These results demonstrate that introducing the local n-frame magnetic vector constraint effectively mitigates heading drift under unobservability conditions. The average RMS heading error decreases from 9.8° to 1.80°, achieving an approximately 70\% improvement in heading estimation accuracy.

\begin{table}[!b]
\caption{RMS Heading Angle Error Statistics (Unit: °)\label{tab:table4}}
\centering
\begin{tabular}{ccccc}
\hline
 & \textbf{aL-1} & \textbf{aL-2} & \textbf{aM-1} & \textbf{aM-2} \\
\hline
OFF	& 3.97 & 8.90 & 6.52  &  19.89 \\
\hline
ON	& 1.82 & 1.55 & 1.94  &  1.90 \\
\hline
\end{tabular}
\end{table}

\subsection{Analysis of the Impact of Magnetic Field Gradients}
\begin{figure*}[!t]
    \centering
    \subfloat[\sffamily\footnotesize aL-1]{\includegraphics[width=0.8\linewidth, trim=80 0 80 0]{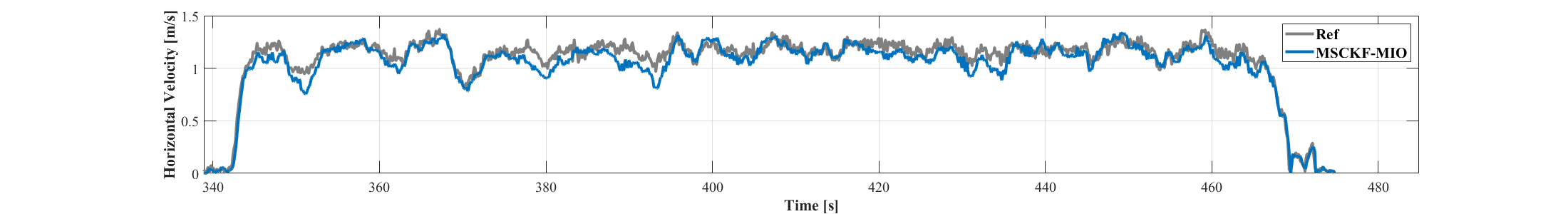}}
    \hfill
    \subfloat[\sffamily\footnotesize aM-1]{\includegraphics[width=0.8\linewidth, trim=80 0 80 0]{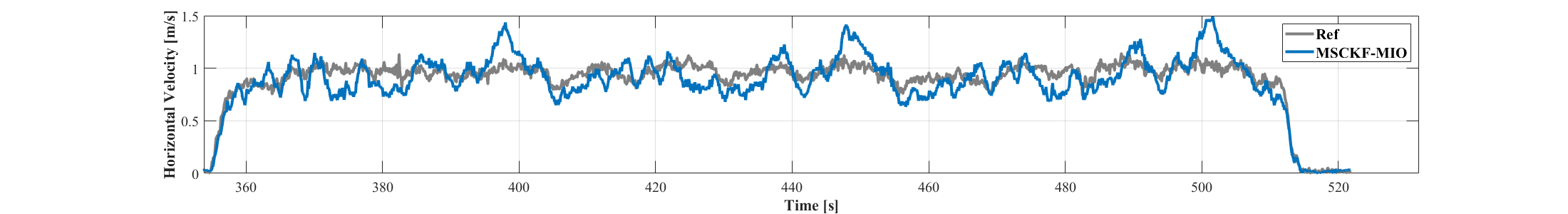}}
    \hfill
    \subfloat[\sffamily\footnotesize aN-1]{\includegraphics[width=0.8\linewidth, trim=80 0 80 0]{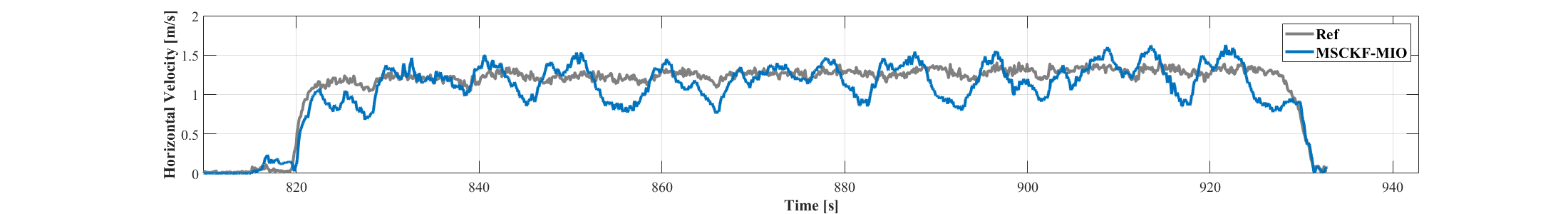}}
              
    \caption{Velocity Estimation Results at Varying Heights in the Same Indoor Environment}
    \label{fig:velo}
\end{figure*}

\begin{figure}[!h]
    \centering
    \includegraphics[width=0.7\linewidth]{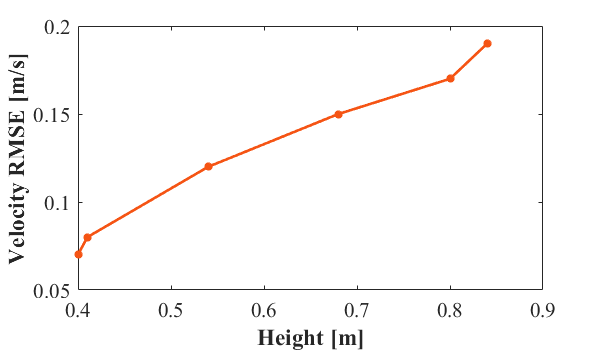}
    \caption{Velocity Error vs. Height}
    \label{fig:height_velo}
\end{figure}

\begin{figure*}[!h]
    \centering
    \subfloat[\sffamily\footnotesize NP-1]{\includegraphics[width=0.4\linewidth, trim=0 80 0 160]{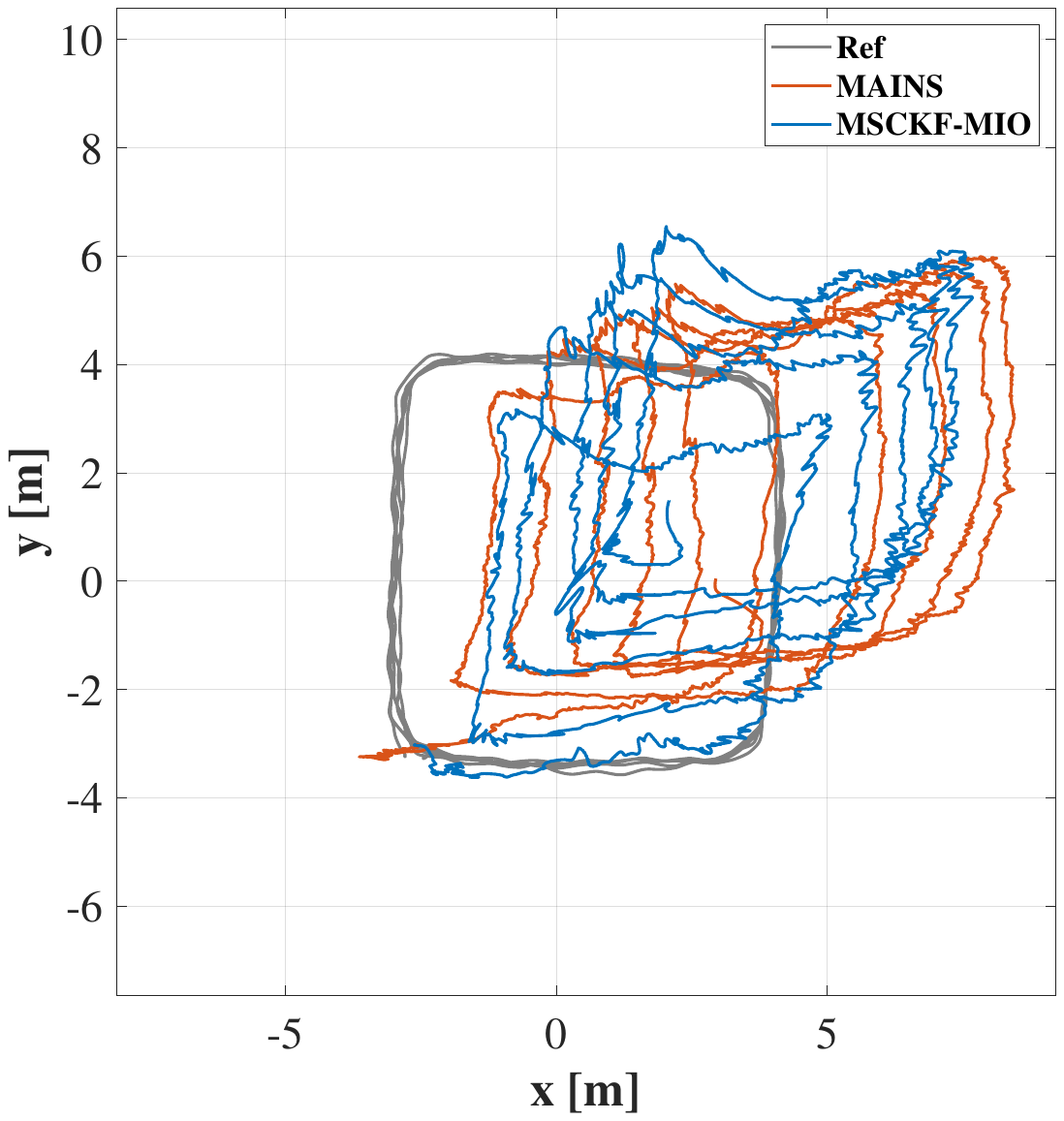}} 
    \subfloat[\sffamily\footnotesize NP-3]{\includegraphics[width=0.4\linewidth, trim=0 80 0 160]{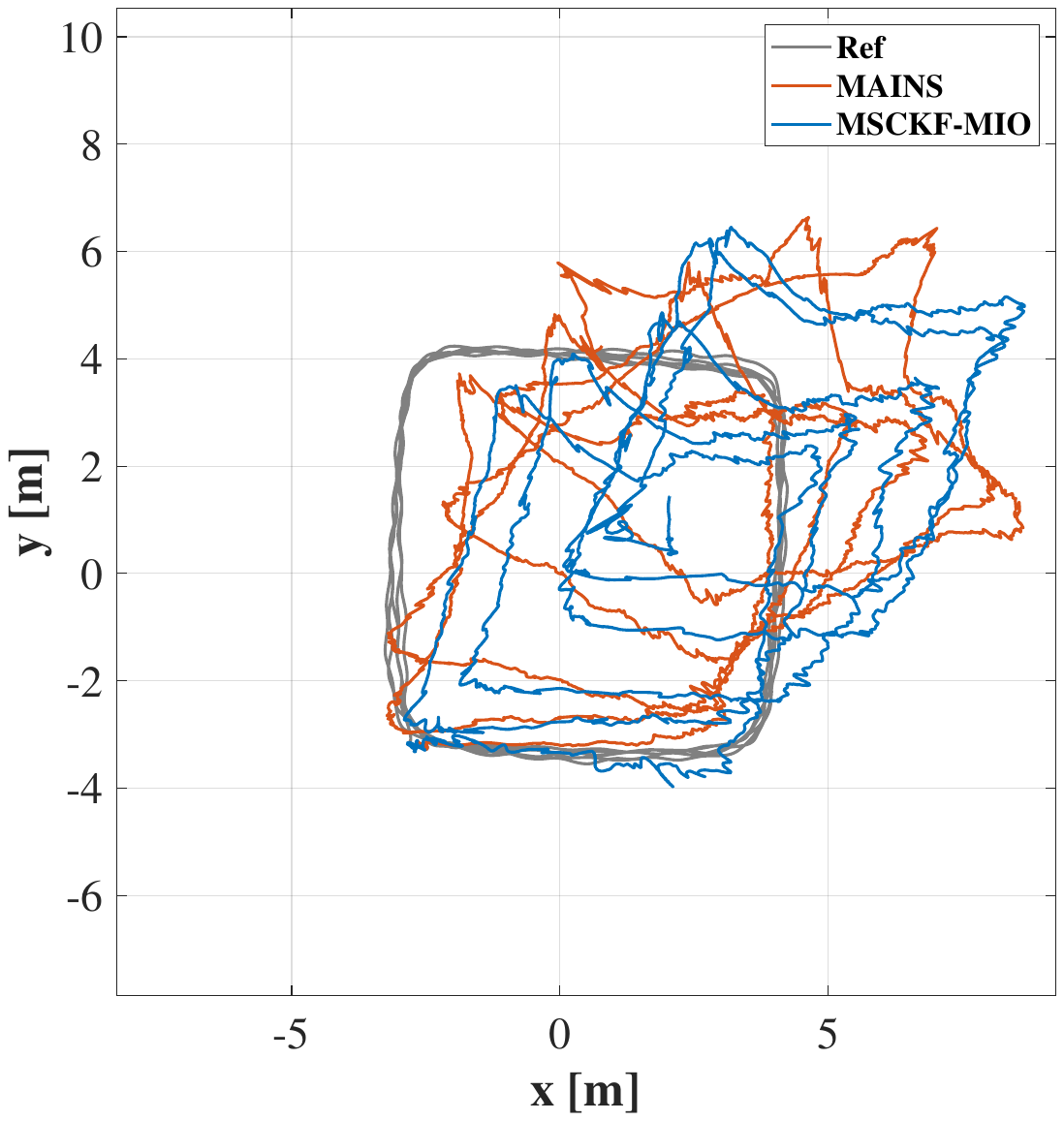}} 
    \hfill
    \subfloat[\sffamily\footnotesize NT-1]{\includegraphics[width=0.4\linewidth, trim=0 80 0 160]{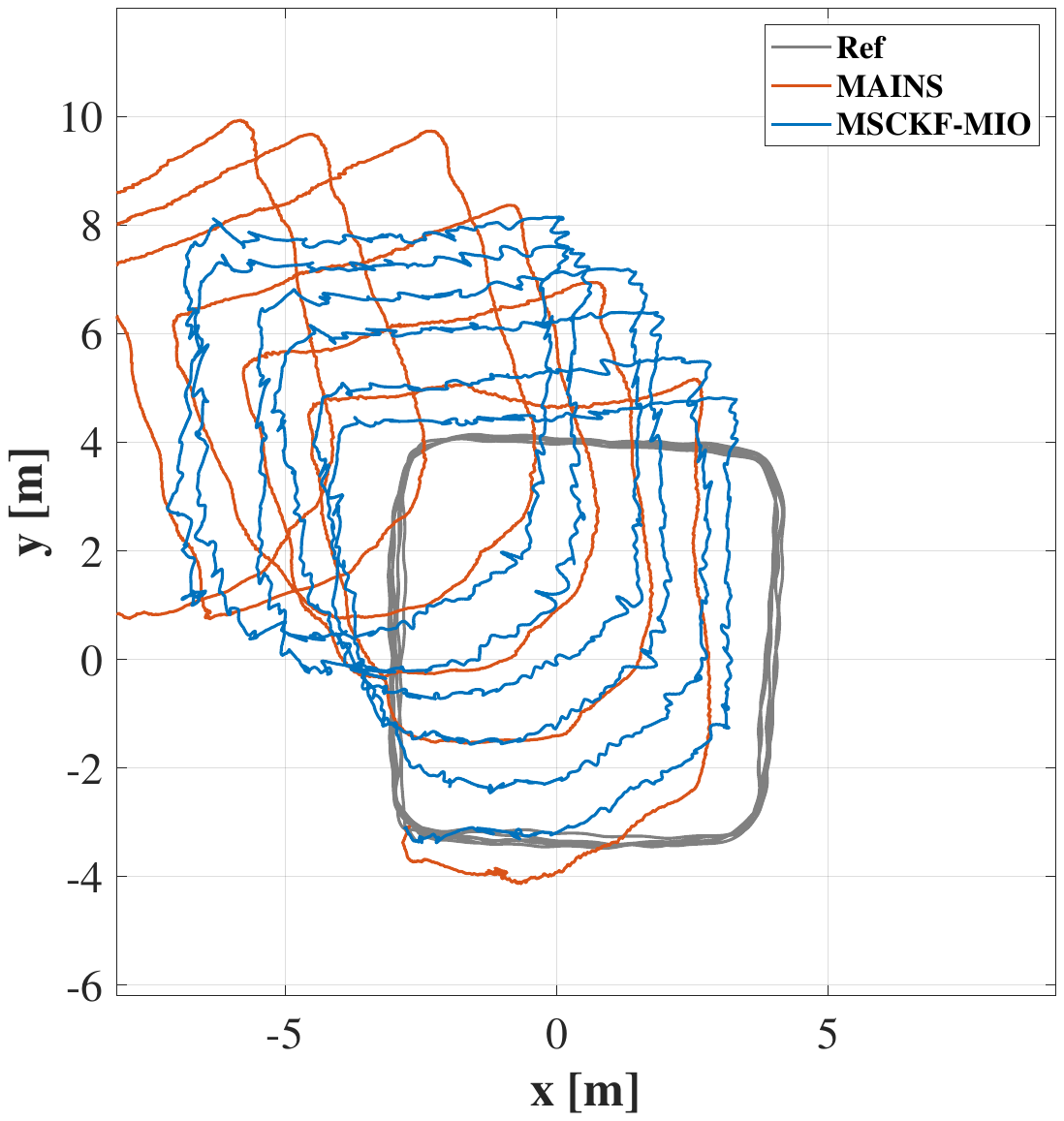}} 
    \subfloat[\sffamily\footnotesize NT-2]{\includegraphics[width=0.4\linewidth, trim=0 80 0 160]{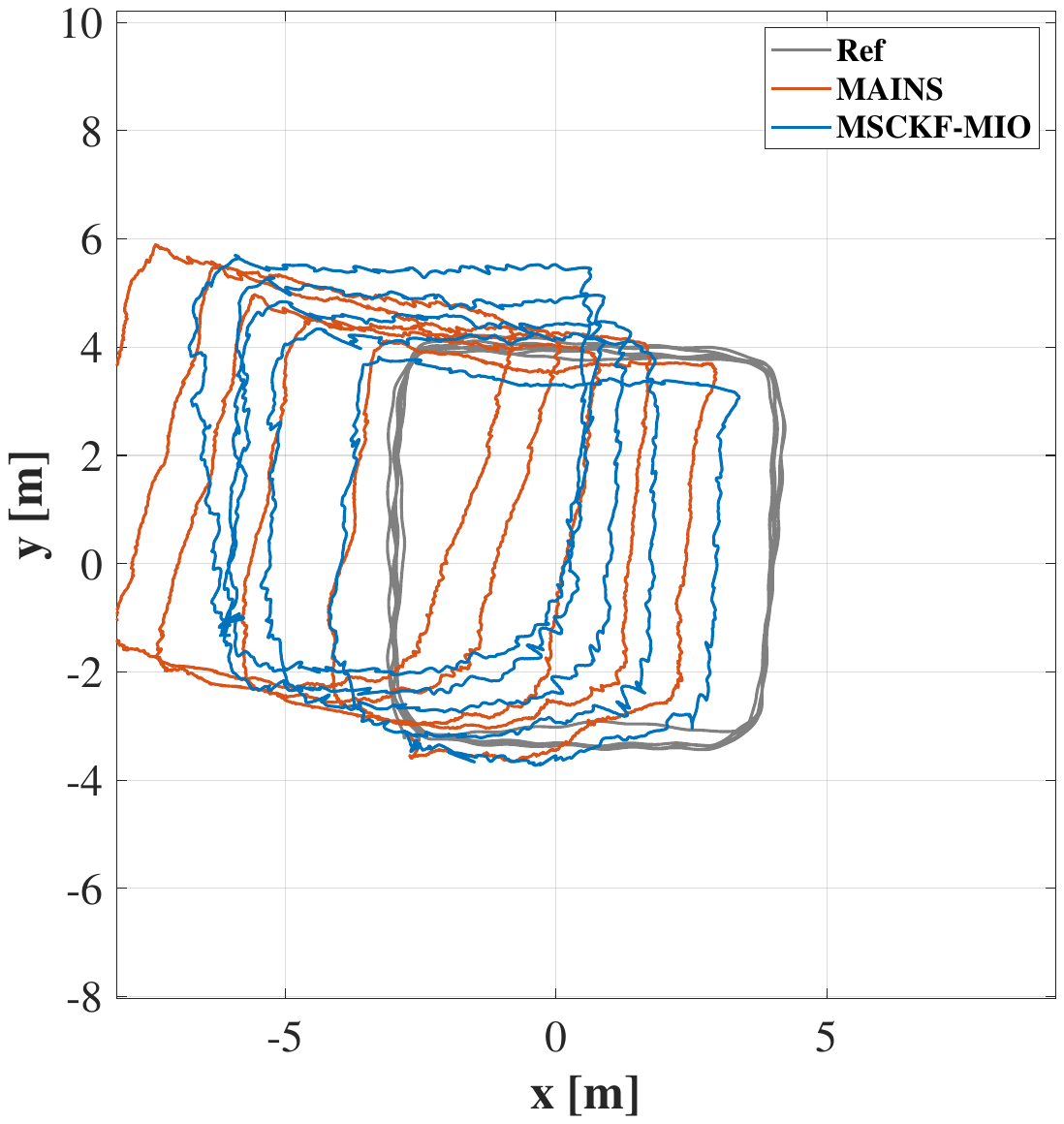}}
    \caption{Localization results for the remaining open datasets}
    \label{fig:other_results}
\end{figure*}

The proposed algorithm utilizes a magnetometer array to measure spatial magnetic gradient variations and constructs constraint information based on the relationship between local gradient changes and the platform's pose. Consequently, significant magnetic gradient variations in the localization environment are critical for the performance of magneto-inertial odometry. Due to the cubic decay of magnetic field strength with distance, this experiment controls the prominence of magnetic gradients by adjusting the sensor platform's height above the indoor ground. Furthermore, the impact of different gradient magnitudes on odometry performance is analyzed. Fig.\ref{fig:velo} display velocity estimation results at varying heights within the same indoor environment. Tab.\ref{tab:table5} lists the corresponding RMS velocity estimation errors, and Fig.\ref{fig:height_velo} shows the RMS error variation with height across the dataset.

\begin{table}[!b]
\caption{RMS Velocity Error Statistics (Unit: m/s)\label{tab:table5}}
\centering
\begin{tabular}{ccccccc}
\hline
 & \textbf{aL-1} & \textbf{aL-2} & \textbf{aM-1} & \textbf{aM-2} & \textbf{aN-1} & \textbf{aN-2} \\
\hline
MAINS	    & 0.27 & 0.26 &	0.21&  0.21	 & 0.34 & 0.35 \\
\hline
MSCEKF-MIO	& 0.07 & 0.08 &	0.12&  0.15  & 0.18 & 0.16\\
\hline
\end{tabular}
\end{table}

The RMS velocity error statistics of the MSCEKF-MIO algorithm in Tab.\ref{tab:table5} demonstrate that magnetic gradient prominence decisively influences algorithmic performance. In the low-height groups (aL-1/aL-2), the mean RMS velocity error is only 0.075 m/s; in the mid-height groups (aM-1/aM-2), the error rises to 0.135 m/s; and in the normal-height groups (aN-1/aN-2), it further increases to 0.17 m/s. This indicates that weakened magnetic gradients introduce significant noise into magnetometer array observations. Nevertheless, MSCEKF-MIO’s error in normal-height groups remains significantly lower than MAINS (RMS error: 0.34–0.35 m/s).

\section{Conclusions and Future Work}
This paper proposes a multi-state constrained magnetic array inertial odometry method (MSCEKF-MIO). The algorithm constructs an environmental magnetic field distribution model through magnetometer array observations and estimates the platform’s absolute velocity by tracking temporal variations in the magnetic model derived from continuous array measurements. These estimates are then used to correct the position, velocity, and attitude obtained via IMU integration. Experimental results demonstrate that the proposed algorithm achieves high localization accuracy in environments with significant magnetic gradient variations. With fewer sensors than MAINS, MSCEKF-MIO exhibits superior performance in both velocity and attitude estimation. As a magneto-inertial odometry framework that requires neither motion assumptions nor prior magnetic fingerprint databases, this work provides a low-power, cost-effective, and highly reliable localization solution for indoor platforms (particularly in tunnels or for pedestrians) under non-cooperative and prior-free conditions. Future work will focus on enhancing the algorithm’s performance in environments with attenuated magnetic gradients. Potential directions include integrating magnetic waveform sequence matching to improve velocity estimation under weak gradient scenarios and leveraging sensors mounted on other body parts to establish collaborative constraints, thereby forming a body sensor network with cooperative localization capabilities.

\begin{table*}[!h]
\caption{Error Statistics of Public Dataset-2\label{tab:table6}}
\centering
\begin{threeparttable}
\begin{tabular}{cccccc}
\hline
\hline
Data sequence & \textbf{NP-1} & \textbf{NP-2} & \textbf{NP-3}& \textbf{NT-1} & \textbf{NT-2} \\
\hline
Trajectory length (m) & 136.23 & 132.17  &	137.76  &	164.62  &	137.87 \\
\hline
Trajectory duration (s) & 177 &	164  &	154  &  185  &	151 \\
\hline
Average height (m) & 0.85  &	0.83  &	0.79  &	0.73  &	0.74 \\
\hline
\rowcolor{gray!20}  \multicolumn{6}{c}{\textbf{MAINS}} \\
\hline
RMS (m)	        & 3.62 &	2.35 &	4.21 &	6.23 &	3.83 \\
\hline
CDF68 (m)	    & 3.77 &	2.55 &	4.37 &	7.25 &	4.63 \\
\hline
RMS Speed (m/s)	& 0.25 &	0.19 &	0.29 &	0.17 &	0.14 \\
\hline
\rowcolor{gray!20}  \multicolumn{6}{c}{\textbf{MSCEKF-MIO}} \\
\hline
RMS (m)	        & 3.56 &	6.68 &	3.58 &	3.71 &	2.87 \\
\hline
CDF68 (m)	    & 4.12 &	7.39 &	4.11 &	4.39 &	3.34 \\
\hline
RMS Speed (m/s)	& 0.17 &	0.26 &	0.22 &	0.10 &	0.09 \\
\hline
\end{tabular}
    \begin{tablenotes}
        \footnotesize
        \item[1] LP: low height and parallel  NP: normal height and parallel NT: normal height and tilted.
    \end{tablenotes}
\end{threeparttable}
\end{table*}

\begin{thebibliography}{1}
\bibliographystyle{IEEEtran}

\bibitem{ref1}
S. A. S. Mohamed, M.-H. Haghbayan, T. Westerlund, J. Heikkonen, H. Tenhunen, and J. Plosila, “A Survey on Odometry for Autonomous Navigation Systems,” IEEE Access, vol. 7, pp. 97466–97486, 2019, doi: 10.1109/ACCESS.2019.2929133.

\bibitem{ref2}
C. Huang, G. Hendeby, H. Fourati, C. Prieur, and I. Skog, “MAINS: A Magnetic Field Aided Inertial Navigation System for Indoor Positioning,” IEEE Sens. J., pp. 1–1, 2024, doi: 10.1109/JSEN.2024.3379932.

\bibitem{ref3}
D. Vissiere, A. Martin, and N. Petit, “Using distributed magnetometers to increase IMU-based velocity estimation into perturbed area,” in 2007 46th IEEE Conference on Decision and Control, Dec. 2007, pp. 4924–4931. doi: 10.1109/CDC.2007.4434809.

\bibitem{ref4}
E. Dorveaux, “Magneto-inertial navigation: principles and application to an indoor pedometer,” Nov. 2011. Accessed: Jan. 12, 2024. [Online].

\bibitem{ref5}
C.-I. Chesneau, M. Hillion, and C. Prieur, “Motion estimation of a rigid body with an EKF using magneto-inertial measurements,” in 2016 International Conference on Indoor Positioning and Indoor Navigation (IPIN), Alcala de Henares, Spain: IEEE, Oct. 2016, pp. 1–6. doi: 10.1109/IPIN.2016.7743702.

\bibitem{ref6}
M. Zmitri, H. Fourati, and C. Prieur, “Improving Inertial Velocity Estimation Through Magnetic Field Gradient-based Extended Kalman Filter,” in 2019 International Conference on Indoor Positioning and Indoor Navigation (IPIN), Sep. 2019, pp. 1–7. doi: 10.1109/IPIN.2019.8911813.

\bibitem{ref7}
I. Skog, G. Hendeby, and F. Gustafsson, “Magnetic Odometry - A Model-Based Approach Using a Sensor Array,” in 2018 21st International Conference on Information Fusion (FUSION), Jul. 2018, pp. 794–798. doi: 10.23919/ICIF.2018.8455430.

\bibitem{ref8}
C. Huang, G. Hendeby, and I. Skog, “A Tightly-Integrated Magnetic-Field aided Inertial Navigation System,” in 2022 25th International Conference on Information Fusion (FUSION), Jul. 2022, pp. 1–8. doi: 10.23919/FUSION49751.2022.9841304.

\bibitem{ref9}
T. Zhang, L. Wei, J. Kuang, H. Tang, and X. Niu, “Mag-ODO: Motion speed estimation for indoor robots based on dual magnetometers,” Measurement, vol. 222, p. 113688, Nov. 2023, doi: 10.1016/j.measurement.2023.113688.

\bibitem{ref10}
W. Zhang, “Ped-Mag-ODO: Indoor Pedestrian Motion Speed Estimation Method Based on Dual Magnetometers,” IEEE Internet Things J., vol. PP, pp. 1–1, Jan. 2024, doi: 10.1109/JIOT.2024.3512504.

\bibitem{ref11}
A. I. Mourikis and S. I. Roumeliotis, “A Multi-State Constraint Kalman Filter for Vision-aided Inertial Navigation,” in Proceedings 2007 IEEE International Conference on Robotics and Automation, Apr. 2007, pp. 3565–3572. doi: 10.1109/ROBOT.2007.364024.

\bibitem{ref12}
J. A. Hesch, D. G. Kottas, S. L. Bowman, and S. I. Roumeliotis, “Consistency Analysis and Improvement of Vision-aided Inertial Navigation,” IEEE Trans. Robot., vol. 30, no. 1, pp. 158–176, Feb. 2014, doi: 0.1109/TRO.2013.2277549.

\bibitem{ref13}
W. Xu, Y. Cai, D. He, J. Lin, and F. Zhang, “FAST-LIO2: Fast Direct LiDAR-inertial Odometry”.

\bibitem{ref14}
Y. Wu, D. Zou, P. Liu, and W. Yu, “Dynamic Magnetometer Calibration and Alignment to Inertial Sensors by Kalman Filtering,” IEEE Trans. Control Syst. Technol., vol. 26, no. 2, pp. 716–723, Mar. 2018, doi: 10.1109/TCST.2017.2670527.

\end{thebibliography}
\end{document}